\documentclass[sn-mathphys]{sn-jnl}

\usepackage[numbers,sort&compress]{natbib}

\usepackage{xcolor}
\usepackage{colortbl}
\definecolor{customgray}{RGB}{240,240,240}
\usepackage{bbding}
\usepackage{pifont}
\usepackage{siunitx}
\usepackage{lmodern}
\jyear{2021}%

\theoremstyle{thmstyleone}%
\theoremstyle{thmstyletwo}%

\theoremstyle{thmstylethree}%

\definecolor{customgray}{RGB}{240,240,240}

\begin{document}

\title[SAM-MI]{SAM-MI: A Mask-Injected Framework for Enhancing Open-Vocabulary Semantic Segmentation with SAM}


\author[1,2]{\fnm{Lin} \sur{Chen}}\email{chenlin2024@ia.ac.cn}

\author[1,2]{\fnm{Yingjian} \sur{Zhu}}\email{zhuyingjian2024@ia.ac.cn}

\author[1,2]{\fnm{Qi} \sur{Yang}}\email{yangqi2021@ia.ac.cn}

\author[3]{\fnm{Xin} \sur{Niu}}\email{2024001016@ruc.edu.cn}

\author*[1]{\fnm{Kun} \sur{Ding}}\email{kun.ding@ia.ac.cn}

\author[1,2]{\fnm{Shiming} \sur{Xiang}}\email{smxiang@nlpr.ia.ac.cn}

\affil[1]{\orgdiv{State Key Laboratory of Multimodal Artificial Intelligence
Systems (MAIS)}, \orgname{Institute of Automation, Chinese Academy of Sciences}, \orgaddress{\city{Beijing} \postcode{100190},  \country{China}}}

\affil[2]{\orgdiv{School of Artificial Intelligence}, \orgname{University of Chinese Academy of Sciences}, \orgaddress{\city{Beijing} \postcode{101408},  \country{China}}}

\affil[3]{\orgdiv{Gaoling School of Artificial Intelligence}, \orgname{Renmin Univeristy of China}, \orgaddress{\city{Beijing} \postcode{100872},  \country{China}}}


\abstract{Open-vocabulary semantic segmentation (OVSS) aims to segment and recognize objects universally. Trained on extensive high-quality segmentation data, the segment anything model (SAM) has demonstrated remarkable universal segmentation capabilities, offering valuable support for OVSS. Although previous methods have made progress in leveraging SAM for OVSS, there are still some challenges: (1) SAM's tendency to over-segment and (2) hard combinations between fixed masks and labels. This paper introduces a novel mask-injected framework, SAM-MI, which effectively integrates SAM with OVSS models to address these challenges. Initially, SAM-MI employs a Text-guided Sparse Point Prompter to sample sparse prompts for SAM instead of previous dense grid-like prompts, thus significantly accelerating the mask generation process. The framework then introduces Shallow Mask Aggregation (SMAgg) to merge partial masks to mitigate the SAM’s over-segmentation issue. Finally, Decoupled Mask Injection (DMI) incorporates SAM-generated masks for guidance at low-frequency and high-frequency separately, rather than directly combining them with labels. Extensive experiments on multiple benchmarks validate the superiority of SAM-MI. Notably, the proposed method achieves a 16.7\% relative improvement in mIoU over Grounded-SAM on the MESS benchmark, along with a 1.6$\times$ speedup.} We hope SAM-MI can serve as an alternative methodology to effectively equip the OVSS model with SAM.




\maketitle

\section{Introduction}
\label{sec:intro}

\begin{figure*}[h!]
\centering
\includegraphics[width=\columnwidth]{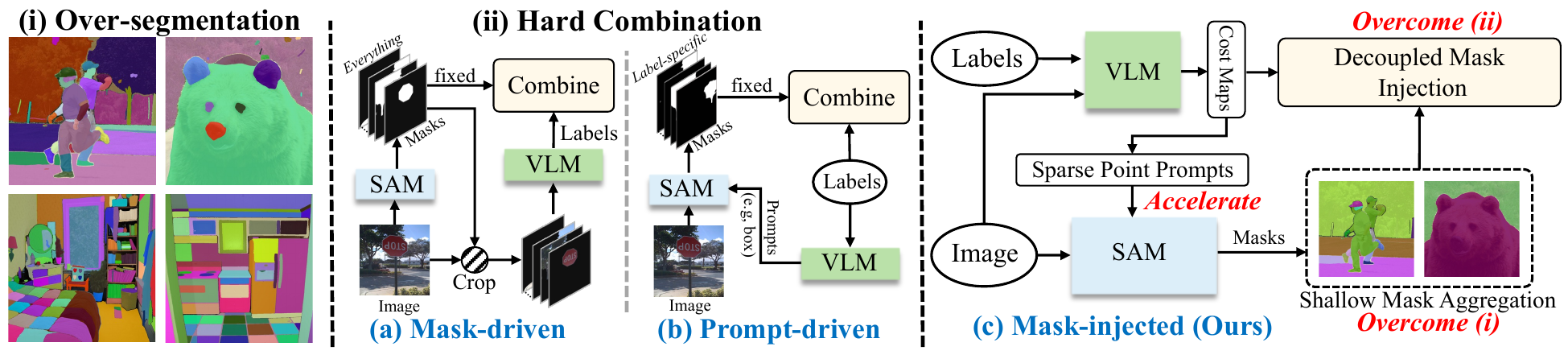}
\caption{Illustration of the challenges addressed by SAM-MI. Previous OVSS methods integrating SAM suffer from the following issues: (i) \textbf{over-segmentation}, SAM may segment numerous out-of-interest patches caused by the edges; (ii) \textbf{hard combination}, previous mask-driven and prompt-driven methods combine fixed masks and labels directly, ignoring the incorrect or low-quality masks. Our proposed mask-injected framework incorporates Shallow Mask Aggregation (SMAgg) to alleviate the over-segmentation issue, and Decoupled Mask Injection (DMI) to avoid the hard combination. Additionally, we accelerate the SAM's masks generation process via learnable sparse point prompts.}
\label{fig:paradigm}
\end{figure*}

Open-vocabulary semantic segmentation (OVSS) aims to achieve image semantic understanding of arbitrary given textual descriptions. Ideally, OVSS models should be able to perform universal segmentation and open-vocabulary recognition. Recently, the OVSS task has witnessed remarkable progress, largely due to the impressive open-vocabulary recognition capabilities of vision-language models (VLMs) that have undergone pre-training on extensive text corpora. Nevertheless, previous approaches~\cite{ding2022decoupling,xu2022simple,liang2023open,dong2023maskclip,xu2023open, xu2023side,yu2024convolutions} mainly rely on the region proposal network (RPN), which is trained within specific domains and lacks the capabilities for universal segmentation. To tackle this limitation, SAM~\cite{kirillov2023segment}, pre-trained on 1.1B masks, is gradually introduced to supply such capability for OVSS models, in consideration of its ability to generate high-quality masks across various domains.

Except for a few methods~\cite{shan2024open, chen2024frozenseg} that only integrate the image features from SAM, prior attempts to leverage SAM in the OVSS task can be categorized into two paradigms: mask-driven and prompt-driven, as shown in Fig.~\ref{fig:paradigm} (a) and (b). The mask-driven paradigm (including SSA~\cite{chen2023semantic}, SAM-CP~\cite{chen2024sam}, SAM-CLIP~\cite{wang2024sam} and OV-SAM~\cite{yuan2024open}) first employs SAM to generate masks within an image as region proposals, and then leverages VLMs to assign text-descriptive labels to them. In contrast, the prompt-driven paradigm (including Grounded-SAM~\cite{ren2024grounded} and SSPrompt~\cite{huang2024learning}) learns prompts (e.g., boxes) from each text description and employs SAM to produce corresponding masks for each prompt. Ultimately, both mask-driven and prompt-driven methodologies establish a mapping between masks and labels, thus forming the final semantic maps.

Although these methods have successfully integrated SAM into the OVSS task, there are still some issues: (\lowercase\expandafter{\romannumeral1}) \textbf{Over-segmentation}, SAM tends to segment regions with colour and boundary clues, while these regions sometimes are out of interest, as shown in Fig.~\ref{fig:paradigm} (\lowercase\expandafter{\romannumeral1}). On the one hand, it poses challenges for VLMs to recognize sub-patches not defined in text labels for mask-driven methods. On the other hand, SAM struggles to segment the precise mask based on text-specific prompts for prompt-driven methods, as there may be multiple masks associated with a single prompt. (\lowercase\expandafter{\romannumeral2}) \textbf{Hard combination}, these methods rigidly associate the fixed masks and labels to derive the final semantic map, as shown in Fig.~\ref{fig:paradigm} (\lowercase\expandafter{\romannumeral2}). This simplistic approach represents a hard combination that overlooks mask accuracy.

To this end, we propose SAM-MI, a novel mask-injected framework designed to equip OVSS with SAM, as shown in Fig.~\ref{fig:paradigm} (c). SAM-MI first extracts coarse-grained pixel-text cost maps from VLMs by measuring cosine similarity between each pixel and text-descriptive labels. To address the issue (\lowercase\expandafter{\romannumeral1}), SAM-MI models the dependencies between different masks from these coarse-grained cost maps and proposes Shallow Mask Aggregation (SMAgg) to reassemble the over-segmented masks. Therefore, SMAgg removes numerous out-of-interest patches, thereby reducing mask redundancy. To address the issue (\lowercase\expandafter{\romannumeral2}), SAM-MI presents the Decoupled Mask Injection (DMI) to improve the contextual information in the pixel-text cost maps through injecting priors from the SAM-generated masks. The injection process is divided into two dimensions: low-frequency and high-frequency, which inject coarse-grained global contextual information and fine-grained local details, respectively. In contrast to mask-driven and prompt-driven approaches, our method utilizes masks as guidance to refine the coarse pixel-text cost maps, enabling the correction of inaccurate or low-quality masks.

In addition, we observe that prior approaches~\cite{chen2023semantic, han2023boosting, chen2024sam, wang2024open}  for mask segmentation heavily rely on dense grid-like point prompts, which leads to excessive computational overhead due to numerous repeated mask decoder forward propagation and non-maximum suppression. SAM-MI proposes the Text-guided Sparse Point Prompter (TSPP) to alleviate this issue by learning text-specific sampling probabilities from pixel-text cost maps. Subsequently, candidate points are sampled based on these probabilities. Compared with the previous dense grid-like sampling approach, TSPP achieves a $96.0\%$ reduction in point prompts on the ADE20K-150~\cite{zhou2019semantic} validation set.

Comprehensive experiments have demonstrated the remarkable performance of our proposed SAM-MI in ADE20K-847~\cite{zhou2019semantic}, PC-459~\cite{mottaghi2014role} and ADE20K-150~\cite{zhou2019semantic}, achieving relative improvements of 4.2\%, 4.2\% and 3.5\% in mIoU compared to SAM-less baselines. Moreover, on the multi-domain semantic segmentation benchmark (MESS)~\cite{MESSBenchmark2023}, SAM-MI also achieves a relative improvement of 16.7\% in mIoU compared with Grounded-SAM~\cite{ren2024grounded}, along with a 1.6$\times$ speedup. Our primary contributions are summarized as follows:

\begin{itemize}
    \item We propose a novel framework for effectively incorporating SAM into open-vocabulary semantic segmentation, showcasing superior performance across various datasets.
    \item To alleviate SAM's tendency to over-segment, we introduce the Shallow Mask Aggregation (SMAgg), which leverages coarse-grained cost maps to assist in merging redundant masks.
    \item To avoid the hard combination between masks and labels and enhance the robustness of incorrect or low-quality masks, we propose the Decoupled Mask Injection (DMI), which decouples low-frequency and high-frequency guidance for cost maps learning.
    \item To accelerate the mask generation process of SAM, we present the Text-guided Sparse Point Prompter (TSPP), which produces sparse prompts instead of traditional dense grid-like points.
\end{itemize}

\section{Related Work}
\label{sec:related}

\subsection{Segment Anything Model.}
The Segment Anything Model (SAM)~\cite{kirillov2023segment} is a vision foundation model for versatile segmentation. Training on extensive and comprehensive masks enables SAM to segment diverse domains and produce detailed masks. Subsequent research on SAM can be categorized into two main groups: enhancements and applications. In terms of enhancements, SSA~\cite{chen2023semantic}, Semantic-SAM~\cite{li2023semantic} and SEEM~\cite{zou2024segment} enable semantic-aware segmentation. HQ-SAM~\cite{ke2024segment} fine-tunes SAM on a high-quality segmentation dataset. FastSAM~\cite{zhao2023fast}, MobileSAM~\cite{zhang2023faster} and EfficientSAM~\cite{xiong2024efficientsam} notably accelerate the inference process of SAM. In terms of applications, SAM has been utilized in various downstream vision tasks~\cite{ji2024segment}, including semantic segmentation~\cite{ren2024grounded}, action recognition~\cite{liu2025balanced}, image captioning~\cite{wang2023caption}, image editing~\cite{gao2023editanything}, anomaly detection~\cite{cao2023segment,li2024clipsam, liu2024deep}, medical segmentation~\cite{zhu2024medical, ma2024segment} etc. This study focuses on its applications, i.e., developing a more powerful OVSS model with SAM.

\subsection{Open-Vocabulary Semantic Segmentation.}
The goal of OVSS is to achieve a pixel-level understanding of arbitrary classes. Most recent OVSS methods are based on large-scale pre-trained foundational models~\cite{radford2021learning, jia2021scaling, li2022blip}. Initially, ZegFormer~\cite{ding2022decoupling} and ZSSeg~\cite{xu2022simple} employ a two-stage approach, utilizing an RPN to generate mask proposals and then performing open-vocabulary recognition for each proposal. Inspired by RegionCLIP~\cite{zhong2022regionclip}, OVSeg~\cite{liang2023open} enhances open-vocabulary recognition by fine-tuning CLIP with region-text pairs. ODISE~\cite{xu2023open} replaces the RPN with a text-to-image diffusion model. 

During the recognition of mask proposals, these two-stage methods depend on multiple forward propagation of VLMs. To avoid this, SAN~\cite{xu2023side} and FC-CLIP~\cite{yu2024convolutions} introduce attention bias and mask-pooling to generate text-space aligned embeddings for each region. Furthermore, EBSeg~\cite{shan2024open} proposes image embedding balancing to alleviate overfitting to training classes, MROVSeg~\cite{zhu2024mrovseg} introduces a multi-resolution training framework. Besides, CAT-Seg~\cite{cho2024cat}, SED~\cite{xie2024sed} and ERR-Seg~\cite{chen2025errseg} model OVSS tasks following the FCN~\cite{long2015fully} methodology, starting from pixel-text cost maps and eliminates the requirement of RPN. In this study, we also model OVSS in the FCN way while incorporating its universal segmentation capabilities.

\subsection{Open-Vocabulary Learning with SAM.}
As a universal segmentor, SAM holds significant importance for open-vocabulary learning. To our knowledge, except for certain methods~\cite{shan2024open, chen2024frozenseg} that only incorporate image features from SAM, previous approaches fall into two primary paradigms: mask-driven and prompt-driven. For mask-driven methods, SSA~\cite{chen2023semantic} uses CLIP to directly label each mask. Moreover, OV-SAM\cite{yuan2024open} and SAM-CLIP\cite{wang2024sam} merge the visual encoders of SAM and CLIP via knowledge distillation for enhanced representations. Sambor~\cite{han2023boosting} employs SAM as an open-set RPN for object detection. SAM-CP~\cite{chen2024sam} introduces composable prompts to group sub-patches belonging to the same instance.

For prompt-driven methods, Grounded-SAM~\cite{ren2024grounded} leverages an offline model (e.g., Grounding DINO~\cite{liu2023grounding}) to generate label-corresponding prompts for SAM. ClipSAM~\cite{li2024clipsam} generates prompts from rough segmentation produced by CLIP. SSPrompt~\cite{huang2024learning} introduces spatial-semantic prompt learning. PerSAM~\cite{zhang2023personalize} learns to prompt from a single image with a reference mask to customize SAM. Automating MedSAM~\cite{gaillochet2024automating} learns prompts from image embedding under tight bounding boxes supervision. In contrast to previous paradigms, we propose a novel mask-injected paradigm, which integrates mask injection to embed low-frequency and high-frequency spatial priors from SAM for OVSS.

\begin{figure*}[t!]
\centering
\includegraphics[width=\columnwidth]{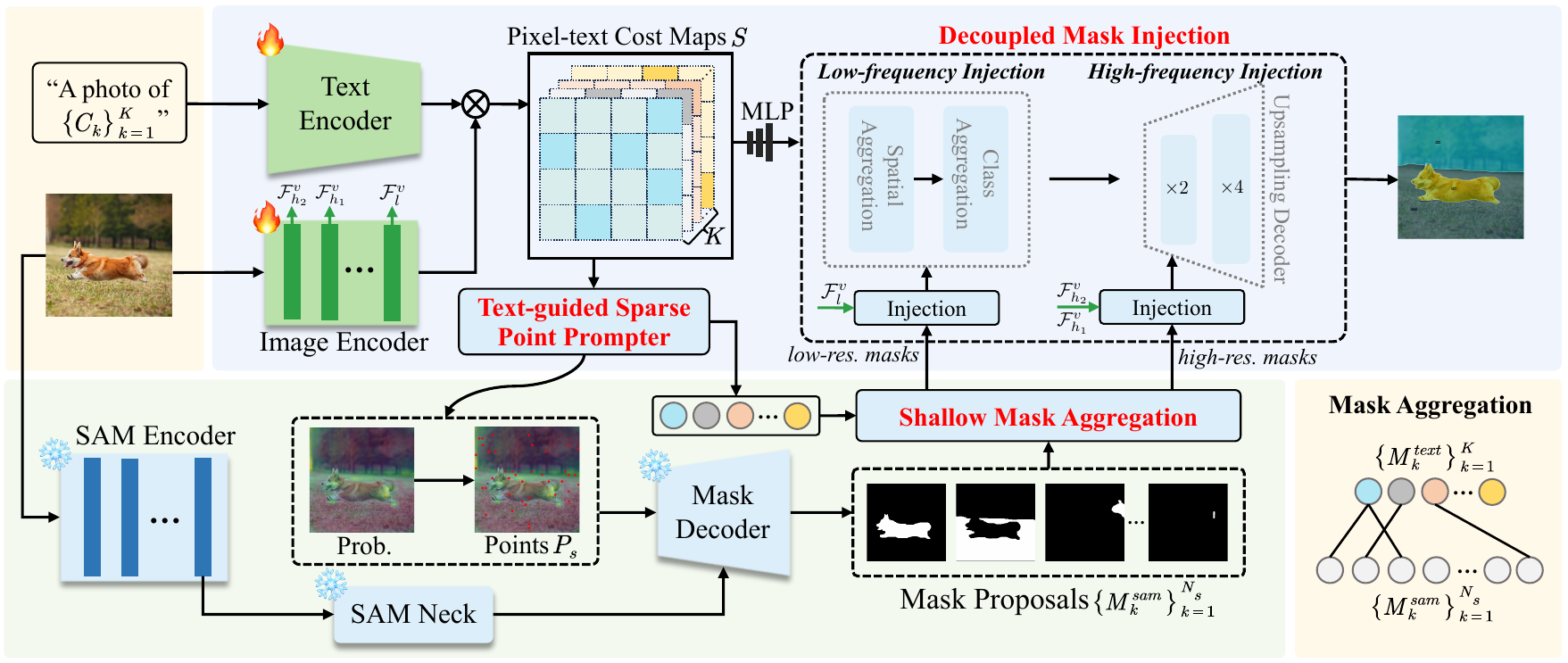}
\caption{The overall architecture of SAM-MI. It begins by utilizing the visual and text encoder from CLIP to produce dense pixel-text cost maps $S$. Subsequently, the Text-guided Sparse Point Prompter is employed to create sparse point prompts for SAM to generate mask proposals $M^{sam}$, which are then partially aggregated by the Shallow Mask Aggregation. Finally, the Decoupled Mask Injection is adopted to inject contextual information from $M^{sam}$ into $S$ in a low-frequency and high-frequency decoupled manner.}
\label{fig:overview}
\end{figure*}

\section{Method}
\label{sec:method}
The architecture of our proposed SAM-MI is depicted in Fig.~\ref{fig:overview}. In this section, we begin with an overview of SAM and CLIP in Sec.~\ref{sec:pre}, followed by a detailed explanation of our Text-guided Sparse Point Prompter in Sec.~\ref{sec:tspp}, Shallow Mask Aggregation in Sec.~\ref{sec:sma}, and Decoupled Mask Injection in Sec.~\ref{sec:dmi}. Finally, we present the training pipeline in Sec.~\ref{sec:training}.

\subsection{Preliminaries}
\label{sec:pre}
\noindent \textbf{SAM.} SAM~\cite{kirillov2023segment} consists of an image encoder, a prompt encoder, and a mask decoder. By utilizing dense grid-like point prompts, SAM can segment all objects within an image. specifically, given an image $I \in \mathbb{R}^{3 \times H \times W}$ and dense point prompts $P_d \in \mathbb{R}^{N\times2}$, the image encoder encodes $I$ into features $\mathcal{F}_{sam} \in \mathbb{R}^{D_{s}\times\frac{H}{16}\times\frac{W}{16}}$, and prompt encoder encodes $P_d$ into prompt tokens $Q_p$. Here, $H \times W$ represents the size of the image $I$, $N$ denotes the prompt count, and $D_s$ is the channel dimension. Subsequently, the mask decoder generates masks and their corresponding confidences by employing cross-attention between $Q_p$ and $\mathcal{F}_{sam}$. In this study, we leverage SAM as a universal RPN for OVSS.

\noindent \textbf{CLIP.} Through image-text contrastive learning, CLIP~\cite{radford2021learning} learns open-world knowledge from language supervision. It involves two well-aligned encoders: an image encoder and a text encoder. Initially, a set of sentences $\mathcal{G}=\{G_k\}_{k=1}^K$ is generated for a given class set $\mathcal{C}=\{C_k\}_{k=1}^K$, where each sentence is constructed by combining prefix prompts with class text, i.e., $G_k$=``a photo of $C_k$". Here, $K$ is the number of classes. Subsequently, the image encoder encodes $I$ into an image embedding $\mathcal{F}_{s} \in \mathbb{R}^{D_c}$, and the text encoder encodes $\mathcal{G}$ into text embeddings $\mathcal{T}\in \mathbb{R}^{K\times D_c}$. The classification score of $I$ for the $k$-th class is then computed as:
\setlength{\abovedisplayskip}{5pt}
\setlength{\belowdisplayskip}{5pt}
\begin{align}
    S_k=\frac{\exp \left( <\mathcal{F} _s,T_k> \right)}{\sum_{i=1}^K{\exp \left( <\mathcal{F} _s,T_i> \right)}},
\end{align}
where $<\cdot,\cdot>$ denotes the cosine similarity.

In this study, following~\cite{rao2022denseclip}, we modify the CLIP's image encoder to generate dense visual embeddings $\mathcal{F}_{d} \in \mathbb{R}^{D_c\times\frac{H}{16}\times\frac{W}{16}}$. By calculating the cosine similarity between $\mathcal{F}_{d}$ and $\mathcal{T}$, we obtain the cost maps $S \in \mathbb{R}^{K\times\frac{H}{16}\times\frac{W}{16}}$, which encapsulate abundant semantic information. Consequently, an MLP layer is adopted to embed $S$ at each class, yielding $S \in \mathbb{R}^{K \times C \times \frac{H}{16}\times\frac{W}{16}}$.

\subsection{Text-guided Sparse Point Prompter}
\label{sec:tspp}

\begin{figure*}
\centering
\includegraphics[width=0.95\columnwidth]{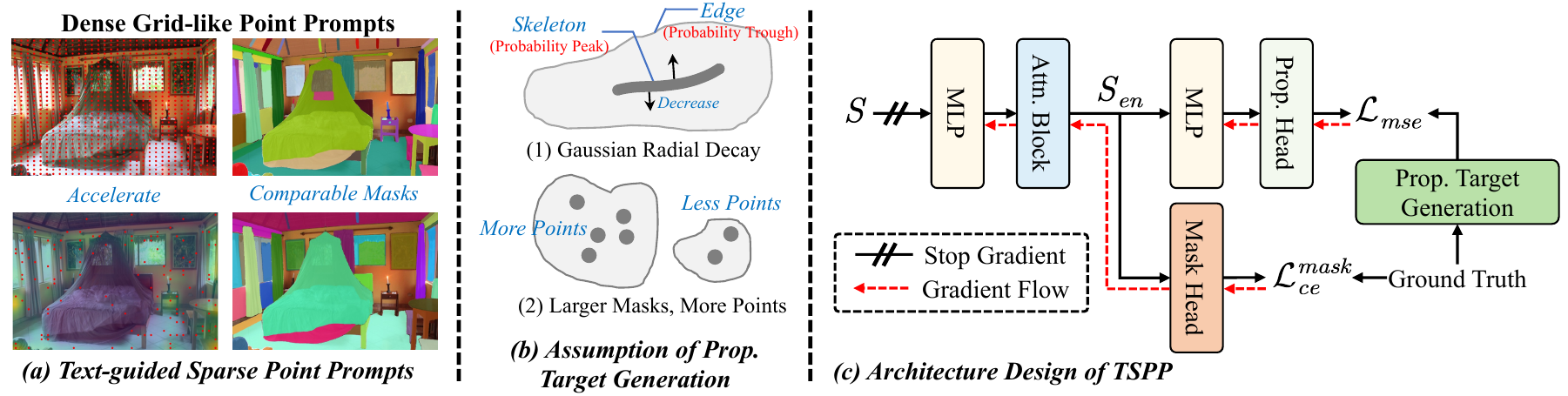}
\caption{Illustration of Text-guided Sparse Point Prompter. (a) Compared with existing dense grid-like point prompts, our Text-guided sparse point prompts achieve comparable masks generated by SAM with much fewer point prompts. (b) The points sampled by TSPP follow two assumptions. (c) The architecture of TSPP consists of two linear projection heads.}
\label{fig:tspp}
\end{figure*}

SAM is a prompt-based interactive segmentation model, and designing prompts for SAM to segment all objects within an image can make it an alternative to RPN. Moreover, compared with RPN trained on specific datasets, SAM possesses universal segmentation capability. However, previous approaches~\cite{chen2023semantic, chen2024sam, han2023boosting, wang2024open} heavily rely on dense grid-like points when utilizing SAM as an RPN. It leads to extensive forward propagation in the SAM's decoder and imposes a substantial non-maximum suppression (NMS) overhead. To alleviate this issue, we introduce the Text-guided Sparse Point Prompter (TSPP), which learns to assign sampling probabilities to the grid-like points guided by text, as shown in Fig.~\ref{fig:tspp}~(a).

\noindent \textbf{Probability Target Generation.}
To train TSPP, we first generate the target sampling probabilities from the ground truth masks $M^{gt} \in \left\{ 0,1 \right\} ^{K\times H\times W}$, represented as $p \in \mathbb{R}^{d_h\times d_w}$. The selected points are denoted as $P_s\in \mathbb{R}^{N_p\times 2}$. Here, $d_h \times d_w$ defines the grid shape, and following previous best practices~\cite{chen2023semantic}, we set $d_h=d_w=32$. $N_p$ is the number of sampled points, and TSPP expects its expectation to be far smaller than sampling all grid points, thus ensuring sparse point prompts, i.e.,
\begin{align}
    E\left( N_p \right) =\sum_{x=1}^{d_h}{\sum_{y=1}^{d_w}{p\left( x,y \right)}}\ll d_h d_w.
\end{align}

As illustrated in Fig.~\ref{fig:tspp}~(b), TSPP assigns probabilities based on two assumptions: (1) points closer to the mask skeleton receive higher probabilities and radial decay in a Gaussian distribution, and (2) the probability of sampling a mask rises with its size. Specifically, $p(x, y)$ can be calculated as:
\setlength{\abovedisplayskip}{5pt}
\setlength{\belowdisplayskip}{5pt}
\begin{align}
p\left( x,y \right) =\frac{\exp \left( -d^2\left( x,y,M_{k}^{gt} \right) /2\sigma _k ^2 \right) P_k}{\sum_{\left( x,y \right) \in M_{k}^{gt}}{\exp \left( -d^2\left( x,y,M_{k}^{gt} \right) /2\sigma _k ^2 \right)}},
\label{tspp1}
\end{align}
where $M_{k}^{gt}$ denotes the mask containing point $(x,y)$, and $p(x,y)$ is set to $0$ if point $(x,y)$ does not belong to any mask. $d(x,y,M_{k}^{gt})$ measures the distance between $(x,y)$ and the skeleton $M_{k}^{gt}$, which can be obtained by distance transformation. $\sigma_k$ stands for the bandwidth parameter of mask $M_{k}^{gt}$, which is estimated by:
\setlength{\abovedisplayskip}{5pt}
\setlength{\belowdisplayskip}{5pt}
\begin{align}
    \sigma _k =\underset{\left( x,y \right) \in M_{k}^{gt}}{\max}d\left( x,y,M_{k}^{gt} \right) /3.
\end{align}
In Eq.~\ref{tspp1}, $P_k$ is the expected number of points to be sampled from $M_{k}^{gt}$ and is determined as follows:
\setlength{\abovedisplayskip}{5pt}
\setlength{\belowdisplayskip}{5pt}
\begin{align}
P_k=\min \left( \lceil sum\left( M_{k}^{gt} \right) /g_p \rceil , m_p \right),
\label{tspp2}
\end{align}
where $g_p$ is a hyperparameter to adjust the density of sampling points, and $m_p$ defines the upper limit of the expected number of points within a mask.

\noindent \textbf{TSPP Training Loss.}
The architecture of TSPP is illustrated in Fig.~\ref{fig:tspp}~(c). It starts with the cost maps $S$, which contain text-related spatial information. Initially, TSPP employs a lightweight MLP layer and attention block to enrich spatial-level and class-level contextual information in $S$, yielding $S_{en}$. Subsequently, probabilities are learned from $S_{en}$ and supervised by target probability maps with a mean-square-error loss $\mathcal{L}_{mse}$. However, experimental findings indicate that training solely with $\mathcal{L}_{mse}$ makes convergence challenging for TSPP. Therefore, we introduce the cross-entropy loss $\mathcal{L}_{ce}^{mask}$ to further enhance the contextual information in $S_{en}$, guided by semantic segmentation labels. Hence, the overall loss for training TSPP is formulated as: $\mathcal{L}_{tspp} = \mathcal{L}_{ce}^{mask} + \lambda_{mse} \mathcal{L}_{mse}$, where $\lambda_{mse}$ is a hyperparameter. Additionally, to prevent overfitting to $\mathcal{L}_{ce}^{mask}$ and $\mathcal{L}_{mse}$ that damages the generalizability of CLIP, we stop its gradient from updating CLIP's visual encoder.

\subsection{Shallow Mask Aggregation}
\label{sec:sma}
By feeding sparse point prompts $P_s$ to SAM, mask proposals $M^{sam}\in \left\{ 0,1 \right\} ^{N_s\times H\times W}$ are generated. The proposals may contain some overly segmented masks that could impede the subsequent injection process. To alleviate this issue, Shallow Mask Aggregation (SMAgg) is proposed to aggregate the masks at a shallower level.

As TSPP also receives the supervision of segmentation labels, it can simultaneously generate text-related masks $M^{text}\in \left\{ 0,1 \right\} ^{K\times \frac{H}{16}\times \frac{W}{16}}$. SMAgg utilizes the coarse-grained spatial information from $M^{text}$ as guidance for the aggregation. SMAgg starts by calculating the matching scores between  $M_{i}^{sam}$ and $M_{j}^{text}$:
\begin{equation}
\mathcal{O} \left( i,j \right) =\frac{ sum\left( M_{i}^{sam}\cap M_{j}^{text} \right) }{ sum\left( M_{i}^{sam} \right) +\varepsilon}.
\end{equation}
Here $\mathcal{O} \in \mathbb{R}^{N_s \times K}$ denotes the matrix of matching scores between $M^{sam}$ and $M^{text}$. $\varepsilon$ is a very small value to avoid denominators with 0. Subsequently, the aggregated mask for the $k$-th class is obtained by jointing those masks in $M^{sam}$ with matching scores to $M_k^{text}$ exceeding $\alpha_{sma}$, i.e., 
\begin{equation}
M_k=\bigcup_{i=1}^{N_s}{\left[ \mathcal{O} \left( i,k \right) >\alpha_{sma} \right] M_{i}^{sam}}.
\end{equation}
Here $[\cdot]$ is a logical expression resulting in 1 if the condition is true and 0 otherwise; $\alpha_{sma}$ is the aggregated threshold discussed in Sec.~\ref{sec:alb-smagg}. Additionally, any instances of $M^{sam}$ not involved in the aggregation are treated individually as a mask and later merged into the output. Ultimately, we obtain $M\in \left\{ 0,1 \right\} ^{N_m\times H\times W}$, where $N_m \leq N_s$.

\begin{figure*}[h]
\centering
\includegraphics[width=0.95\columnwidth]{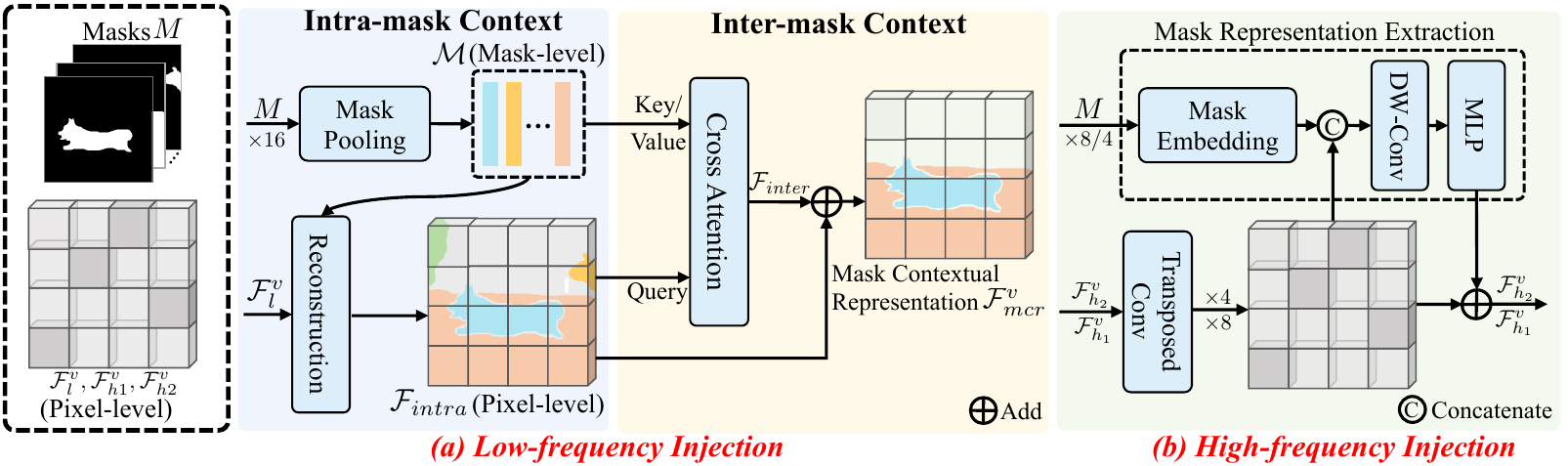}
\caption{Design of Decoupled Mask Injection. (a) The low-frequency injection models intra-mask context by mask-pooling and inter-mask context by cross attention in low-resolution feature map $\mathcal{F}_{l}^v$. (b) The high-frequency injection extracts and injects high-frequency representations from the high-resolution feature maps $\mathcal{F}_{h1}^v$ and $F_{h2}^v$.}
\label{fig:dmi}
\end{figure*}
\subsection{Decoupled Mask Injection}
\label{sec:dmi}
Prior methods~\cite{wang2024sam, yuan2024open, ren2024grounded} associate each mask with its label in a hard manner directly. This hard manner gives no opportunity to refine the mask, leading to sub-optimal semantic maps. In contrast, SAM-MI develops the Decoupled Mask Injection (DMI) to inject $M$ into $S$. Our architecture adopts the cost aggregation framework~\cite{cho2024cat} for OVSS, which comprises a spatial and class-level aggregation module for low-frequency refinement and an upsampling decoder for high-frequency details reconstruction. Correspondingly, our injection procedure is decoupled into low-frequency injection and high-frequency injection.

\noindent \textbf{Low-frequency Injection.}
Low-frequency injection focuses on enhancing coarse-grained global contextual information in the aggregation module using low-frequency masks $M$. Cost aggregation framework leverages the last-layer output of CLIP image encoder $\mathcal{F}_{l}^{v} \in \mathbb{R}^{D_v \times \frac{H}{16} \times \frac{W}{16}}$ as a similarity measurement between different pixels, where $D_v$ is the channel dimension. We further enhance $\mathcal{F}_{l}^{v}$ to obtain the mask contextual representation $\mathcal{F}_{mcr}^{v}$ through the modelling of both intra-mask and inter-mask contexts, as illustrated in Fig.~\ref{fig:dmi}~(a).

For intra-mask contexts, mask-pooling is first applied to $\mathcal{F}^v_l \in \mathbb{R}^{D_v \times \frac{H}{16} \times \frac{W}{16}}$ to create mask-level representations for each mask, denotes as $\mathcal{M} \in \mathbb{R}^{N_m\times D_v}$.
\begin{align}
    \mathcal{M} _k=\frac{1}{sum\left( M_k \right)}\sum_{\left( x,y \right) \in M_k}{\mathcal{F}_l^v}\left( x,y \right),
\end{align}
Subsequently, the intra-mask context $\mathcal{F}_{intra}$ is reconstructed by $M$ and $\mathcal{M}$ as, 
{\setlength{\abovedisplayskip}{5pt}
\setlength{\belowdisplayskip}{5pt}
\begin{equation}
    \mathcal{F}_{intra} \left( x,y \right) =\mathcal{F}_l^{v}\left( x,y \right) +\sum_{k=1}^{N_m}{\left[ \left( x,y \right) \in M_k \right] \cdot \mathcal{M} _k}.
\end{equation}}

\noindent The inter-mask context then further captures the dependencies between different masks through a cross-attention mechanism between $\mathcal{F}_{intra}$ and $\mathcal{M}$. Specifically, $\mathcal{F}_{intra}$ serves as the query and $\mathcal{M}$ serves as the key and value:
\begin{align}
    \mathcal{F}_{inter} = \mathrm{Attention}(Q=\mathcal{F}_{intra},\  KV=\mathcal{M}).
\end{align}
In this way, a pixel's embedding is reorganized based on its similarity with $\mathcal{M}$, denoted as $\mathcal{F}_{inter}$. The mask contextual representation $\mathcal{F}_{mcr}^{v} \in \mathbb{R}^{\frac{H}{16} \times \frac{W}{16} \times D_v}$ is then obtained by adding $\mathcal{F}_{inter}$ and $\mathcal{F}_{intra}$.

\noindent \textbf{High-frequency Injection.}
High-frequency injection aims to leverage the precise edges in $M$ to enhance local details in the upsampling decoder. Cost aggregation has shown the feasibility of using the shallow outputs $\mathcal{F}_{h}^v = \mathcal{F}_{h1}^v$ or $\mathcal{F}_{h2}^v$ of the CLIP image encoder as guidance for high-frequency reconstruction. Therefore, we proceed to incorporate the local details from $M$ into $\mathcal{F}_{h}^v$, as shown in Fig.~\ref{fig:dmi}~(b).

The core idea of our high-frequency injection is to extract high-frequency features from the masks $M$ to enhance $\mathcal{F}_{h}^v$. Initially, $M$ is embedded into a dense embedding $M_{emd}$ of identical size as $\mathcal{F}_{h}$, and concatenated with it:
\begin{align}
    M_{emd} = \mathrm{concat}[\mathcal{F}_{h}^v,\  \mathrm{proj}(M)],
\end{align}
where $\mathrm{concat}[\cdot, \cdot]$ represents the function for concatenating features along the channel dimension, and $\mathrm{proj}$ is an MLP projector to embed $M$ into the latent space. Subsequently, a depthwise convolution and an MLP layer are utilized to extract the information from $M_{emd}$. Finally, this extracted information is added to the original $\mathcal{F}_{h}^v$:
 \begin{align}
    \mathcal{F}_{h}^v \leftarrow \mathcal{F}_{h}^v + \gamma \cdot \text{DW-Conv}(\text{MLP}(M_{emd})),
\end{align}
where $\gamma$ is learnable weights and initialized to 1. As a result, more detailed guidance for upsampling is obtained, replacing subsequent computations.

\subsection{Two-stage Training}
\label{sec:training}
Initially, the pseudo label generated by TSPP is random, leading to low-quality aggregated masks for SMAgg. Therefore, directly training SAM-MI with randomly initialized TSPP hinders the convergence of SAM-MI. Therefore, we adopt a two-stage training approach: (1) only training the parameters of TSPP using the $\mathcal{L}_{tspp}$ loss while keeping other parameters frozen, and (2) training the complete SAM-MI model while updating the TSPP parameters with a reduced learning rate. The overall loss function utilized during the second stage is defined as: $\mathcal{L} =\mathcal{L} _{ce}+\lambda_{tspp} \mathcal{L} _{tspp}$, where $\lambda_{tspp}$ is a hyperparameter to adjust TSPP training during stage 2. After stage 1, TSPP was well-initialized, thus a small weight is sufficient and we set it to $0.1$.

\section{Experiment}
\label{sec:experiment}
\subsection{Experimental Setup}
\label{sec:setup}
\noindent \textbf{Datasets and Evaluation Metrics.} 
SAM-MI is trained on COCO-Stuff~\cite{caesar2018coco}, which comprises 118,287 training images with 171 different classes. Following prior OVSS methodologies~\cite{xu2023side, xie2024sed, shan2024open, cho2024cat, zhu2024mrovseg}, evaluations are conducted on five well-designed benchmarks, covering three datasets: PASCAL VOC~\cite{everingham2010pascal} (PAS-20), PASCAL-Context~\cite{mottaghi2014role} (PC-459/PC-59) and ADE20K~\cite{zhou2019semantic} (A-847/A-150). To further validate cross-domain robustness, we extend evaluation to the MESS (Multi-domain Evaluation of Semantic Segmentation) benchmark~\cite{MESSBenchmark2023} containing 22 specialized datasets across five domains: general (e.g., BDD100K~\cite{yu2020bdd100kdiversedrivingdataset}), earth monitoring (e.g., iSAID~\cite{waqas2019isaid}), medical sciences (e.g., CryoNuSeg~\cite{mahbod2021cryonuseg}), engineering (e.g., DeepCrack~\cite{zou2018deepcrack}), and agriculture and biology (e.g., CWFID~\cite{haug2015crop}).

\noindent \textbf{Implementation Details.}
We adopt OpenAI pre-trained CLIP~\cite{cherti2023reproducible} as our VLM, and its visual encoder architecture is ViT~\cite{dosovitskiy2020vit}. Following the methodology of CAT-Seg~\cite{cho2024cat}, both CLIP's visual and text encoders are fine-tuned. To generate high-quality masks, we adopt frozen SAM-Huge, with the point prompts sampled from $32 \times 32$ grids. The image is resized to $384\times 384$ for CLIP and $1024 \times 1024$ for SAM. The hyperparamters are set as $g_p=5$, $m_p=10$, $\lambda_{mse}=0.5$, $\lambda_{tspp}=0.1$ and $\alpha_{sma}=0.50$. Our model is trained using the AdamW~\cite{loshchilov2017decoupled} optimizer. The initial learning rate is $2\times10^{-4}$, and the fine-tuning learning rate of CLIP is set to $2\times10^{-6}$. For a fair comparison, we train our model for $80$K iterations with a batch size of $4$, following prior methods~\cite{xie2024sed, cho2024cat}. $4$ NVIDIA RTX 3090 GPUs are used for training. All latency is reported on the A-150 validation set using a single NVIDIA RTX 3090 GPU.

\vspace{-10pt}
\begin{table*}[h]
    \caption{The comparison of the multi-domain semantic segmentation benchmark (MESS)~\cite{MESSBenchmark2023} shows the performance of various models. The terms LB and UB represent the lower bound and upper bound, respectively. The results indicate that SAM-MI exhibits strong generalization performance and achieves impressive results in diverse fields such as earth monitoring.
    }
    \label{tab:mess}
    \begin{center}
    \resizebox{\textwidth}{!}{
    \begin{tabular}{l|cccccc|ccccc|cccc|cccc|ccc|c}
    \toprule
 & \multicolumn{6}{c|}{General} & \multicolumn{5}{c|}{Earth Monitoring} & \multicolumn{4}{c|}{Medical Sciences} & \multicolumn{4}{c|}{Engineering} & \multicolumn{3}{c|}{Agri. and Biology} & \\
 & \rotatebox[origin=l]{90}{BDD100K} & \rotatebox[origin=l]{90}{Dark Zurich} & \rotatebox[origin=l]{90}{MHP v1} & \rotatebox[origin=l]{90}{FoodSeg103} & \rotatebox[origin=l]{90}{ATLANTIS} & \rotatebox[origin=l]{90}{DRAM} & \rotatebox[origin=l]{90}{iSAID} & \rotatebox[origin=l]{90}{ISPRS Pots.} & \rotatebox[origin=l]{90}{WorldFloods} & \rotatebox[origin=l]{90}{FloodNet} & \rotatebox[origin=l]{90}{UAVid} & \rotatebox[origin=l]{90}{Kvasir-Inst.} & \rotatebox[origin=l]{90}{CHASE DB1} & \rotatebox[origin=l]{90}{CryoNuSeg} & \rotatebox[origin=l]{90}{PAXRay-4} & \rotatebox[origin=l]{90}{Corrosion CS} & \rotatebox[origin=l]{90}{DeepCrack} & \rotatebox[origin=l]{90}{PST900} & \rotatebox[origin=l]{90}{ZeroWaste-f} & \rotatebox[origin=l]{90}{SUIM} & \rotatebox[origin=l]{90}{CUB-200} & \rotatebox[origin=l]{90}{CWFID} & \rotatebox[origin=l]{90}{Mean} \\
\midrule
\textit{Random (LB)} & \phantom{0}\textit{1.48} & \phantom{0}\textit{1.31} & \phantom{0}\textit{1.27} & \phantom{0}\textit{0.23} & \phantom{0}\textit{0.56} & \phantom{0}\textit{2.16} & \phantom{0}\textit{0.56} & \phantom{0}\textit{8.02} & \textit{18.43} & \phantom{0}\textit{3.39} & \phantom{0}\textit{5.18} & \textit{27.99} & \textit{27.25} & \textit{31.25} & \textit{31.53} & \textit{9.30} & \textit{26.52} & \phantom{0}\textit{4.52} & \phantom{0}\textit{6.49} & \phantom{0}\textit{5.30} & \phantom{0}\textit{0.06} & \textit{13.08} & \textit{10.27} \\
\textit{Best supervised (UB)} & \textit{44.80} & \textit{63.90} & \textit{50.00} & \textit{45.10} & \textit{42.22} & \textit{45.71} & \textit{65.30} & \textit{87.56} & \textit{92.71} & \textit{82.22} & \textit{67.80} & \textit{93.70} & \textit{97.05} & \textit{73.45} & \textit{93.77} & \textit{49.92} & \textit{85.90} & \textit{82.30} & \textit{52.50} & \textit{74.00} & \textit{84.60} & \textit{87.23} & \textit{70.99} \\
\midrule
ZSSeg~\cite{xu2022simple} & 32.36 & 16.86 & \phantom{0}7.08 & \phantom{0}8.17 & 22.19 & 33.19 & \phantom{0}3.80 & 11.57 & 23.25 & 20.98 & 30.27 & 46.93 & 37.0 & \textbf{38.70} & \textbf{44.66} & \phantom{0}3.06 & 25.39 & 18.76 & \phantom{0}8.78 & \underline{30.16} & \phantom{0}4.35 & 32.46 & 22.73 \\
ZegFormer~\cite{ding2022decoupling} & 14.14 & \phantom{0}4.52 & \phantom{0}4.33 & 10.01 & 18.98 & 29.45 & \phantom{0}2.68 & 14.04 & 25.93 & 22.74 & 20.84 & 27.39 & 12.47 & 11.94 & 18.09 & \phantom{0}4.78 & 29.77 & 19.63 & 17.52 & 28.28 & \underline{16.80} & 32.26 & 17.57 \\
SAN~\cite{xu2023side} & 37.40 & 24.35 & \phantom{0}8.87 & \underline{19.27} & \underline{36.51} & 49.68 & \phantom{0}4.77 & \underline{37.56} & 31.75 & \textbf{37.44} & \underline{41.65} & \textbf{69.88} & 17.85 & 11.95 & 19.73 & \phantom{0}3.13 & \underline{50.27} & \underline{19.67} & \textbf{21.27} & 22.64 & \textbf{16.91} & \phantom{0}5.67 & 26.74 \\
X-Decoder~\cite{zou2023generalized} & 47.29 & 24.16 & \phantom{0}3.54 & \phantom{0}2.61 & 27.51 & 26.95 & \phantom{0}2.43 & 31.47 & 26.23 & \phantom{0}8.83 & 25.65 & 55.77 & 10.16 & 11.94 & 15.23 & \phantom{0}1.72 & 24.65 & 19.44 & 15.44 & 24.75 & \phantom{0}0.51 & 29.25 & 19.80 \\
OpenSeeD~\cite{zhang2023simple} & \textbf{47.95} & \underline{28.13} & \phantom{0}2.06 & \phantom{0}9.00 & 18.55 & 29.23 & \phantom{0}1.45 & 31.07 & 30.11 & 23.14 & 39.78 & \underline{59.69} & \textbf{46.68} & 33.76 & 37.64 & \underline{13.38} & 47.84 & \phantom{0}2.50 & \phantom{0}2.28 & 19.45 & \phantom{0}0.13 & 11.47 & 24.33 \\
Grounded-SAM~\cite{ren2024grounded} & 41.58 & 20.91 & \textbf{29.38} & 10.48 & 17.33 & \underline{57.38} & \underline{12.22} & 26.68 & \underline{33.41} & 19.19 & 38.34 & 46.82 & 23.56 & \underline{38.06} & \underline{41.07} & \textbf{20.88} & \textbf{59.02} & \textbf{21.39} & 16.74 & 14.13 & \phantom{0}0.43 & \underline{38.41} & \underline{28.52} \\
\rowcolor{customgray} SAM-MI (ours) & \underline{47.76} &\textbf{29.95} &\underline{25.22} &\textbf{24.97} &\textbf{40.90} &\textbf{65.65} &\textbf{20.30} &\textbf{44.41} &\textbf{43.73} &\underline{36.27} &\textbf{43.58} &49.44 &\underline{27.51} &13.83 &37.54 &13.24 &31.67 &19.49 &\underline{18.18} &\textbf{43.12} &10.57 &\textbf{44.74} & \textbf{33.28} \\
        \bottomrule
    \end{tabular}
    }
    \end{center}
\end{table*}

\vspace{-22pt}
\begin{table*}[h]
\centering
\caption{Comparison on five standard OVSS benchmarks. We also report their VLM architectures, training datasets, and whether to leverage SAM.
}
\label{tab:sota}
\resizebox{\textwidth}{!}{
\begin{tabular}{l|c|c|c|ccccc}
\toprule
Method                              & Architecture & SAM & Training Datasets & A-847 & PC-459 & A-150 & PC-59 & PAS-20 \\ \midrule
OVSeg~\cite{liang2023open}          & ViT-B/16     & \ding{55}    & COCO-Stuff       & 7.1   & 11.0   & 24.8  & 53.3  & 92.6   \\
DeOP~\cite{han2023open}          &ViT-B/16    & \ding{55}    &COCO-Stuff-156 &7.1   &9.4   &22.9 &48.8  &91.7 \\
SAN~\cite{xu2023side}               & ViT-B/16  & \ding{55}    & COCO-Stuff        & 10.1  & 12.6   & 27.5  & 53.8  & 94.0   \\
SAM-CLIP~\cite{wang2024sam}         & ViT-B/16  & \checkmark   & Merged-41M        & -     & -      & 17.1  & 29.2  & -      \\
SCAN~\cite{liu2024open}  & ViT-B/16  & \ding{55}   & COCO-Stuff   &10.8  &13.2  &30.8  &58.4  &97.0 \\
EBSeg~\cite{shan2024open}           & ViT-B/16  & \checkmark   & COCO-Stuff     & 11.1 & 17.3 & 30.0 & 56.7 & 94.6 \\
SED~\cite{xie2024sed}               & ConvNeXt-B  & \ding{55} & COCO-Stuff        & 11.4 & 18.6 & 31.6 & 57.3 & 94.4 \\
CAT-Seg~\cite{cho2024cat}           & ViT-B/16  & \ding{55}   & COCO-Stuff        & 12.0  & 19.0   & 31.8  & 57.5  & 94.6   \\
\rowcolor{customgray} SAM-MI (ours) & ViT-B/16   & \checkmark  & COCO-Stuff & \textbf{12.5} & \textbf{19.8}  & \textbf{32.9}  & \textbf{58.4} & \textbf{95.2} \\ \midrule
ODISE~\cite{xu2023open}             & ViT-L/14  & \ding{55}   & COCO-Panoptic     & 11.1  & 14.5   & 29.9  & 57.3  & -      \\
FC-CLIP~\cite{yu2024convolutions}   & ConvNeXt-L  & \ding{55} & COCO-Panoptic     & 14.8  & 18.2   & 34.1  & 58.4  & 95.4   \\
SAM-CP~\cite{chen2024sam}           & ConvNeXt-L & \checkmark  & COCO-Panoptic     & -     & -      & 31.8  & -     & -      \\
MAFT~\cite{jiao2023maft}      & ConvNeXt-L  & \ding{55}  & COCO-Stuff       & 14.6  & 19.5   & 32.6  & 52.2  & 93.0   \\ 
MAFT+~\cite{jiao2024maftp}      & ConvNeXt-L  & \ding{55}  & COCO-Stuff       & 15.1  & 21.6   & 36.1  & 59.4  & 96.5   \\ 
Unpair-Seg~\cite{wang2024open}      & ConvNeXt-L  & \checkmark  & Merged-130K       & 14.6  & 19.5   & 32.6  & 52.2  & 93.0   \\ 
FrozenSeg~\cite{chen2024frozenseg} & ConvNeXt-L & \checkmark &COCO-Panoptic & - & 19.7 & 34.4 & 59.9 & - \\ 
CAT-Seg~\cite{cho2024cat}           & ViT-L/14   & \ding{55}  & COCO-Stuff  & 16.0  & 23.8   & 37.9  & 63.3  & 97.0   \\
OVSNet~\cite{niu2025eov}           & ConvNeXt-L   & \ding{55}  & COCO-Panoptic  & 16.2  & 23.5   & 37.1  & 62.0  & 96.9   \\
MaskCLIP++~\cite{zeng2024maskclippp}           & ViT-L/14   & \ding{55}  & COCO-Stuff  & \textbf{16.8}  & 23.9   & 38.2  & 62.5  & 96.8   \\
\rowcolor{customgray} SAM-MI (ours) & ViT-L/14   & \checkmark  & COCO-Stuff & 16.3   & \textbf{24.4}     &  \textbf{38.6} & \textbf{63.9}      & \textbf{97.2} \\ \bottomrule
\end{tabular}
}
\end{table*}

\subsection{Main Results}
\textbf{Comparisons with state-of-the-art methods.}
The cross-domain evaluation on MESS (Table~\ref{tab:mess}) reveals SAM-MI's superior performance across all domains, with an average relative improvement of 16.7\% in mIoU. It demonstrated that SAM-MI exhibits strong generalization capabilities and performs effectively across various domains. Notably, in the earth monitoring domain, Grounded-SAM achieves an average mIoU of 25.97 across five datasets, while SAM-MI reaches 37.66 mIoU, showcasing a remarkable relative improvement of 45.0\% over Grounded-SAM. This highlights the advantage of our proposed mask-injected paradigm over the prompt-driven approach of Grounded-SAM. Our paradigm leverages SAM-generated masks as guidance, effectively avoiding the over-segmentation and hard combination issues inherent in SAM. In contrast, Grounded-SAM first generates bounding boxes and then uses them to prompt SAM for mask generation, which suffers from these issues, leading to suboptimal performance.

Table~\ref{tab:sota} presents the mIoU results of various methods on five standard OVSS benchmarks. Most methods utilize the COCO-Stuff or COCO-Panoptic as the training dataset. All methods employ CLIP~\cite{radford2021learning} as the VLM, albeit with different VLM architectures. When employing ViT-B/16 VLM architecture, SAM-MI demonstrates remarkable performance on A-847, PC-459, and A-150, surpassing CAT-Seg~\cite{cho2024cat} by a relative improvement of 4.2\%, 4.2\% and 3.5\% in mIoU, respectively. This highlights SAM's ability to address the granularity limitations of middle-scale VLMs effectively. Conversely, when employing larger VLMs that produce more precise masks, the enhancements are moderated, yielding relative improvements of 1.9\%, 2.5\%, and 1.8\% on A-847, PC-459, and A-150, respectively.

\begin{figure}[htbp]
    \centering
    \includegraphics[width=\linewidth]{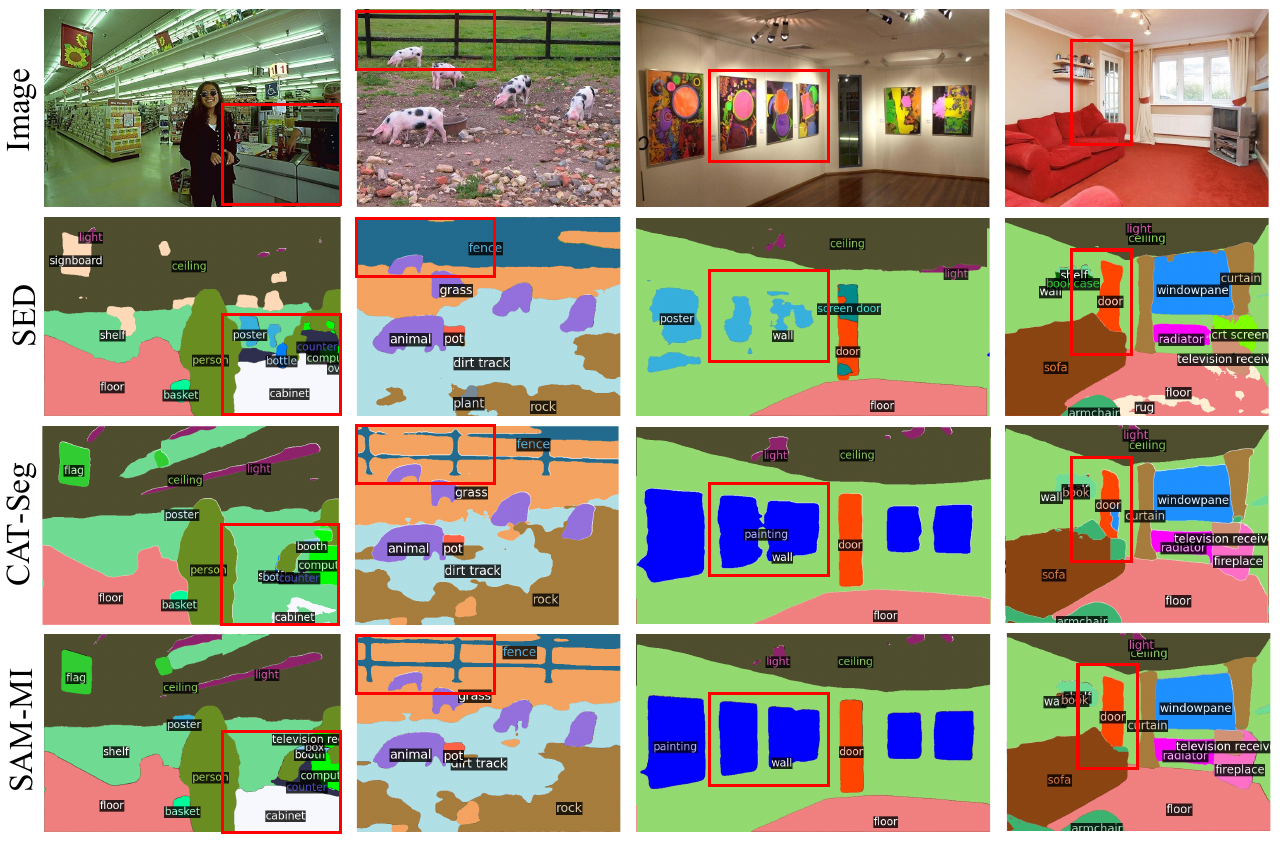}
    \caption{Qualitative comparison with SED~\cite{xie2024sed} and CAT-Seg~\cite{cho2024cat} on ADE20K-150 validation set. SAM-MI is capable of producing more accurate masks.}
    \label{fig:quality}
    \vspace{-20pt}
\end{figure}

\noindent \textbf{Qualitative results.}
Fig.~\ref{fig:quality} presents a qualitative comparison among SED~\cite{xie2024sed}, CAT-Seg~\cite{cho2024cat} and our proposed SAM-MI on the ADE20K-150 validation set. The results demonstrate that SAM-MI, leveraging the universal segmentation capability of SAM, can produce more precise masks compared to SED and CAT-Seg.

\begin{table}[htbp]
    \centering
    \caption{Ablation study for the modules introduced in SAM-MI. Each module is removed individually to analyze its impact.}
    \label{tab:component}
    \begin{tabular}{l|c|ccccc}
    \toprule
    Module  &Latency (ms) $\downarrow$     & A-847 & PC-459 & A-150 & PC-59 & PAS-20 \\ \midrule
    SAM-MI      & 950    & \textbf{12.5}  & \textbf{19.8}   & \textbf{32.9}  & 58.4  & \textbf{95.2}   \\
    $w/o$ DMI   & -  & 12.0  & 19.0  & 31.8  & 57.5  & 94.6  \\
    $w/o$ SMAgg & -   & 11.2  & 18.0   & 30.7  & 56.1  & 92.8    \\
    $w/o$ TSPP  & 5458  & \textbf{12.5}  & \textbf{19.8}   & 32.8  & \textbf{58.6}  & \textbf{95.2}    \\ \bottomrule
    \end{tabular}
    \vspace{-20pt}
\end{table}

\subsection{Ablation Study and Analysis}
\textbf{Component analysis.}
Table~\ref{tab:component} presents the outcomes of individual removal of our proposed modules: Decoupled Mask Injection (DMI), Text-guided Sparse Point Prompter (TSPP), and Shallow Mask Aggregation (SMAgg). It reveals that omitting DMI from SAM-MI, which performs prior injection from SAM, leads to significant performance degradation (e.g., 1.1 mIoU decrease on A-150), thereby validating the efficacy of our mask-injected paradigm. Removing SMAgg forces DMI to process over-segmented masks, resulting in even even more significant degradation across all benchmarks, such as a 2.2 mIoU reduction on A-150. This suggests that fragmented masks substantially impair the injection process. Moreover, removing SMAgg and DMI significantly impacts performance highlights our SAM-MI's efforts to address the issues outlined in Fig.~\ref{fig:paradigm}, which lead mask-driven and prompt-driven methods to struggle with SAM-generated low-quality masks. Substituting TSPP with a uniform grid sampling strategy increases inference latency by $475\%$ while yielding negligible performance gains.

\noindent \textbf{Ablation study for DMI.}
Table~\ref{tab:dmi} quantitatively analyzes the impact of our DMI module. The performance on each dataset is enhanced by low-frequency injection alone (e.g., a 0.6 mIoU improvement on PC-459). Furthermore, adding high-frequency injection further boosts the performance (e.g., 0.3 mIoU improvement on PC-459). This indicates that low-frequency injection alone is insufficient to capture the fine details of the mask generated by SAM, which can be compensated with high-frequency injection. 

\begin{table}[h]
    \centering
    \vspace{-10pt}
    \caption{Ablation Study for DMI. Low and high represent low-frequency injection and high-frequency injection in DMI, respectively.}
    \label{tab:dmi}
    \begin{tabular}{ cc *{5}{S[table-format=2.1]} }
    \toprule
    \multicolumn{2}{c}{$w/o$ DMI} & 
    \multicolumn{5}{c}{Dataset} \\
    \cmidrule(lr){1-2} \cmidrule(l){3-7}
    {Low} & {High} & {A-847} & {PC-459} & {A-150} & {PC-59} & {PAS-20} \\ 
    \midrule
    &  
    & 12.0 & 19.0 & 31.8 & 57.5 & 94.6 \\
    
    $\checkmark$ & 
    & 12.4 & 19.6 & 32.5 & 58.1 & \textbf{95.3} \\ 
    
     & $\checkmark$
    & 12.4 & 19.5 & 32.7 & 58.2 & 95.2 \\ 
    
    $\checkmark$ & $\checkmark$ 
    & \textbf{12.5} & \textbf{19.8} & \textbf{32.9} & \textbf{58.4} & 95.2 \\ 
    \bottomrule
    \end{tabular}
\end{table}

\begin{figure}[htbp]
    \centering
    \vspace{-5pt}
    \includegraphics[width=0.82\linewidth]{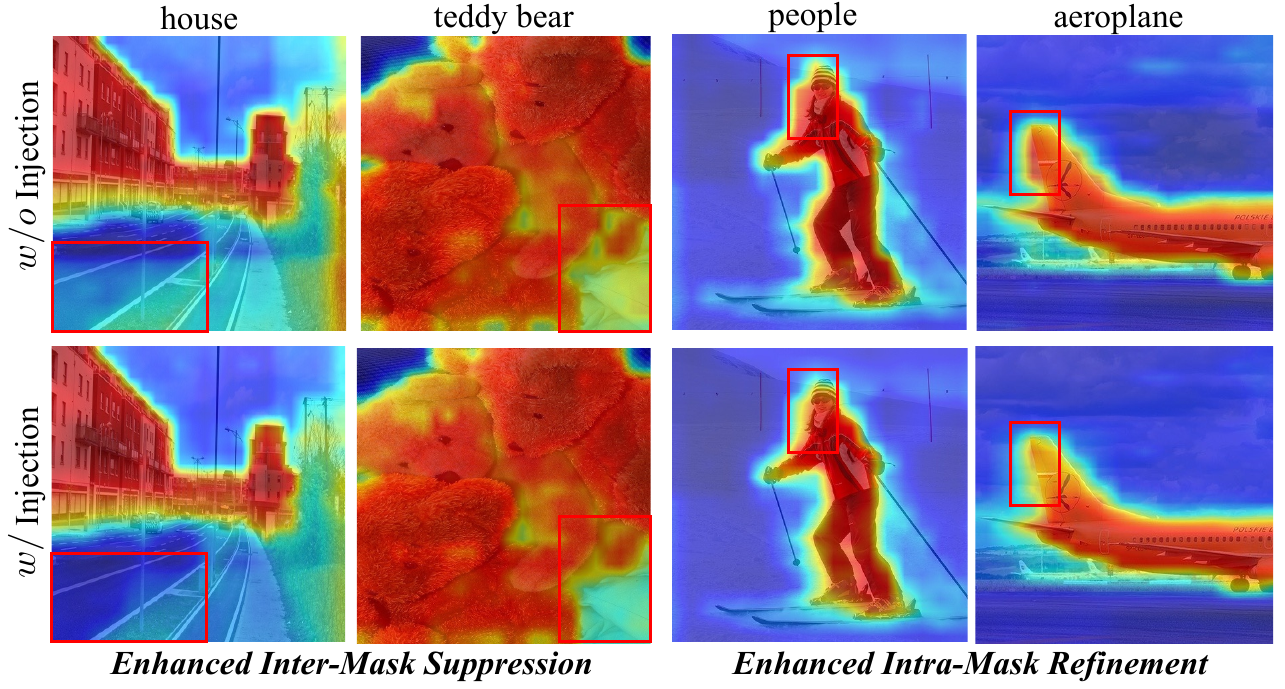}
    \caption{Impact of Low-frequency Injection to prediction logits. Red indicates a strong response, while blue indicates a weak response. It reveals the effectiveness of inter-mask context modeling and intra-mask context modeling in low-frequency injection, i.e., enhancing inter-mask suppression while maintaining intra-mask refinement.}
    \label{fig:low-inj}
\end{figure}

\begin{figure}[h]
    \centering
    \includegraphics[width=0.5\linewidth]{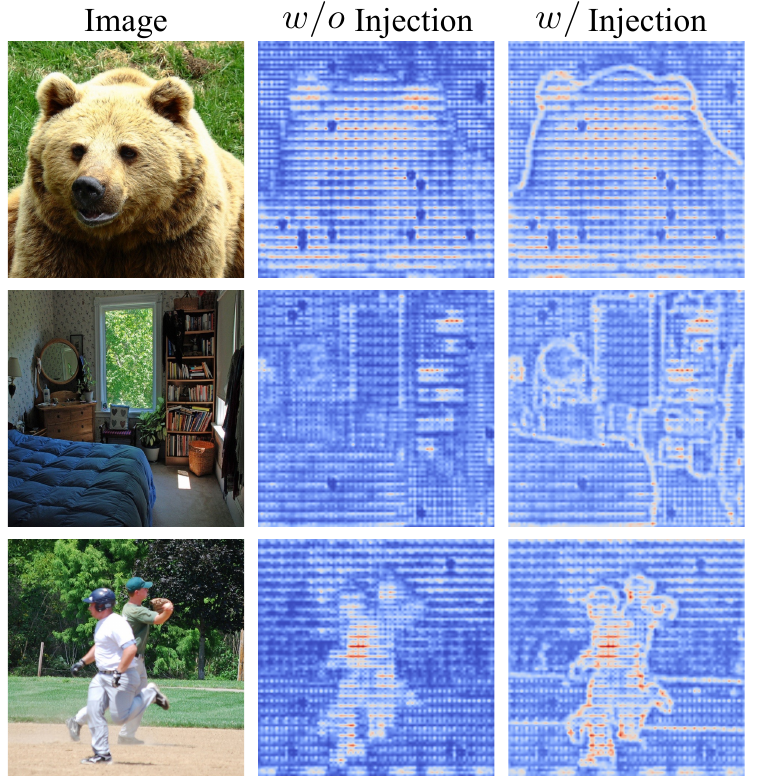}
    \caption{Impact of high-frequency injection. It shows that high-frequency injection can enhance the high-frequency edge information in the feature map.}
    \label{fig:high-inj}
\end{figure}

Fig.~\ref{fig:low-inj} visualizes the cost map after aggregation and before upsampling, revealing the mechanism behind low-frequency injection. It demonstrates two key effects:  (1) Enhanced inter-mask suppression, where identifying one class leads to greater suppression of the other class, and (2) enhanced intra-mask refinement, where the mask for the identified class becomes clearer. These effects align with the concepts of inter-mask context modeling and intra-mask context modeling introduced during low-frequency injection. Fig.~\ref{fig:high-inj} visualizes the feature map after high-frequency injection, showing that the boundaries in the feature map become clearer.

\begin{figure}[h]
    \centering
    \includegraphics[width=\linewidth]{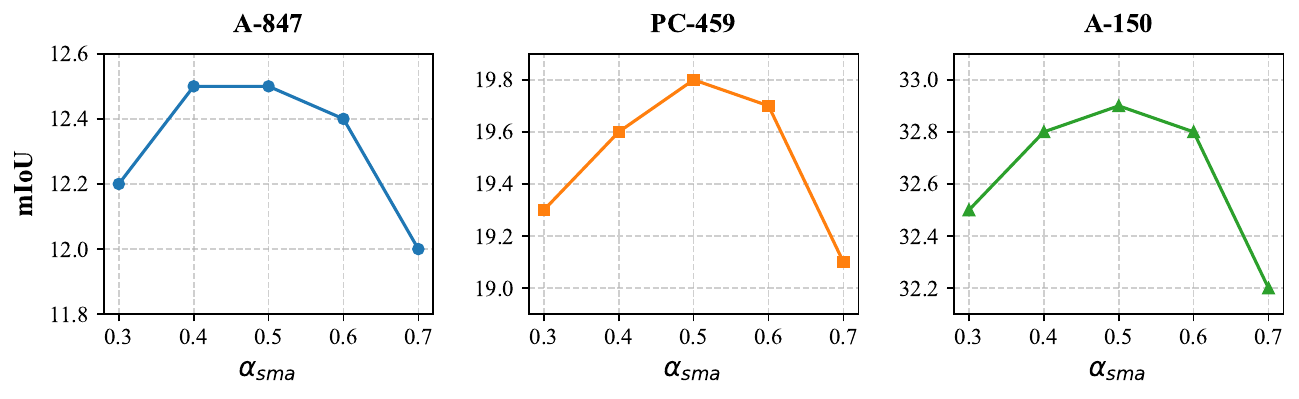}
    \caption{Analysis on aggregated threshold in SMAgg. $\alpha_{sma}$ is the hyperparameter for SMAgg to regulate the combination of over-segmented masks.}
    \label{fig:smagg}
\end{figure}

\noindent \textbf{Analysis on aggregated threshold in SMAgg.}
\label{sec:alb-smagg}
In Fig.~\ref{fig:smagg}, various $\alpha_{sma}$ values are utilized to assess the influence of SMAgg. Lower $\alpha_{sma}$ values lead to the aggregation of more over-segmented masks, and higher $\alpha_{sma}$ values result in fewer combinations. Experimental results indicate that $\alpha_{sma} \in [0.4, 0.6]$ yields stable performance, with optimal performance achieved at $\alpha_{sma}=0.5$. However, performance degrades significantly when $\alpha_{sma}$ deviates beyond this range (e.g., dropping to 0.3 or 0.7). Specifically, the mIoU for A-847 decreases by 0.5 when $\alpha_{sma}$ increases from 0.5 to 0.7, and by 0.3 when $\alpha_{sma}$ decreases to 0.3. This demonstrates the importance of mask aggregation quality. As the pixel-text cost map is coarse-grained, a lower $\alpha_{sma}$ may combine masks that do not belong to the same object, while a higher $\alpha_{sma}$ may ignore some masks that belong to the same object.

\begin{figure*}[t]
\centering
\includegraphics[width=\columnwidth]{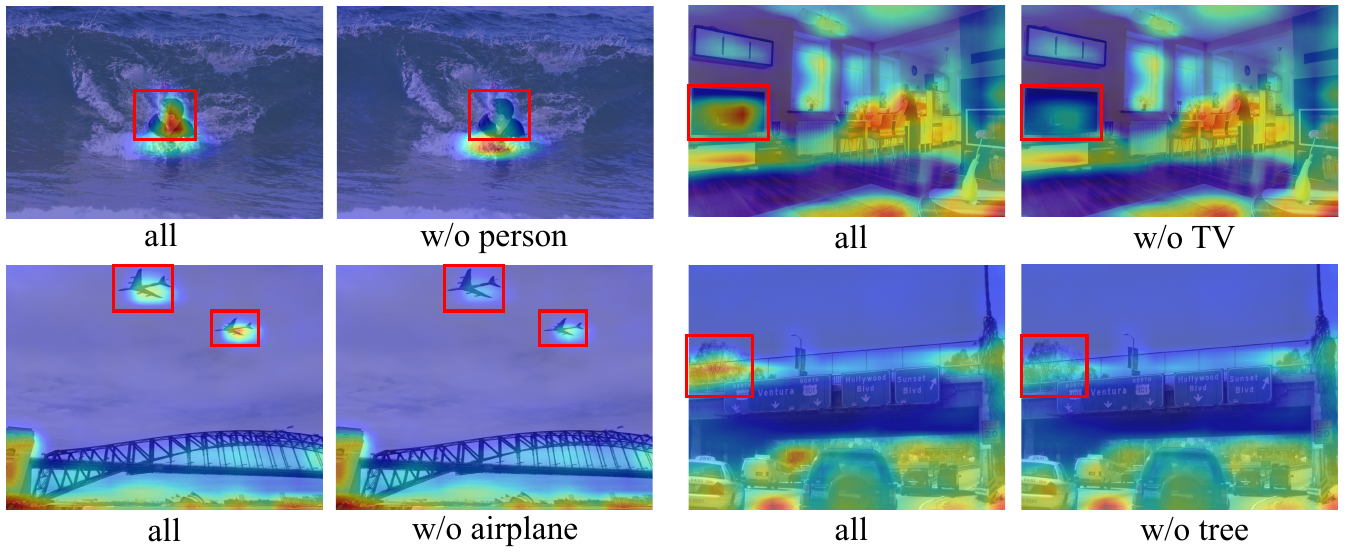}
\caption{Visual examples of probability maps generated by TSPP on the COCO-Stuff validation set. Learning from pixel-text cost maps, TSPP demonstrates its capability in text-guided probability generation. Meanwhile, it assigns higher probabilities to small objects (e.g., person and airplane) than to large objects (e.g., ocean, sky) to ensure sparse sampling.}
\label{fig:tspp_visual}
\end{figure*}

\begin{table}[t]
    \centering
    \vspace{-7pt}
    \caption{Performance Comparison of Sampling Strategies. Uniform Grid denotes dense grid-like sampling methods. Random Sampling represents choosing random points in the $32 \times 32$ grid.}
    \label{tab:tspp}
    \begin{tabular}{@{}lcccccc@{}}
    \toprule
    \multirow{2}{*}{Sampling Strategy} & 
    \multicolumn{3}{c}{A-150} & 
    \multicolumn{3}{c}{PC-59} \\ 
    \cmidrule(lr){2-4} \cmidrule(lr){5-7}
    & Points$\downarrow$ & Latency$\downarrow$ & mIoU$\uparrow$ & Points$\downarrow$ & Latency$\downarrow$ & mIoU$\uparrow$ \\ 
    \midrule
    Uniform Grid ($32 \times 32$) & 1024   & 5967   & 32.8      & 1024     & 5458  & \textbf{58.6}     \\ 
    Random Sampling               & 64     & 1741   & 32.6      & 64   & 1579   & 58.2     \\ 
    TSPP (ours)     & \textbf{41}     & \textbf{950}    & \textbf{32.9}    & \textbf{37}   & \textbf{847}    & 58.4         \\ 
    \bottomrule
    \end{tabular}
\end{table}

\begin{table}[t]
    \centering
    \vspace{-7pt}
    \caption{Average number of points sampled per image by TSPP across various datasets.}
    \label{tab:tspp_adapt_points}
    \begin{tabular}{l|ccccc}
    \toprule
  Dataset        & A-847 & PC-459 & A-150 & PC-59 & PAS-20 \\ \midrule
  Average Points & 47 & 39 & 41 & 37 & 32 \\ \bottomrule
    \end{tabular}
\end{table}

\noindent \textbf{Analysis on TSPP.}
Fig.~\ref{fig:tspp_visual} visualizes some sampling probability maps generated by TSPP. It shows that sampling probabilities are higher for small objects than for large objects, thus ensuring that objects of different sizes have similar sampling points. For instance, in the first image, the person and skateboard have a high sampling probability, while the probabilities for the ocean are lower. It ensures that small masks can be sampled and large masks do not require so many points. Additionally, the probability generation of TSPP demonstrates text-guided characteristics. Removal of the person, tree, aeroplane and TV results in decreased sampling probability at the respective positions. 

Table~\ref{tab:tspp} quantitatively analyzes sampling efficiency. Our TSPP achieves comparable performance to $32 \times 32$ grid sampling with only an average of 41 sampled points on A-150, representing a 96\% reduction. This efficiency translates to significant computational gains, yielding inference speedups of around 6$\times$ on the A-150 and PC-59 datasets, compared to uniform grid sampling. When matching point budgets, TSPP outperforms random sampling by 0.3 and 0.2 mIoU improvement on A-150 and PC-59, suggesting better sampling efficiency. 

As shown in Table~\ref{tab:tspp_adapt_points}, the average number of points sampled by TSPP shows a increasing trend as the number of categories in the dataset increases. This phenomenon, along with the text-guided sampling probabilities demonstrated in Fig.~\ref{fig:tspp_visual}, confirms TSPP's adaptability to different datasets. This capability stems from its ability to learn from pixel-text cost maps. Different datasets generate cost maps associated with diverse classes, resulting in adjustments to both the distribution and the number of sampled points.

\noindent \textbf{Hyperparameter analysis of TSPP.}
Table~\ref{tab:tspp_hyper} presents the influence of hyperparameters $m_p$ and $g_p$, which control the density of sampled points by TSPP. Increasing $m_p$ or decreasing $g_p$ results in a higher density. We observe that increasing the sampling density over our baseline configuration ($m_p=10$, $g_p=5$) has a negligible effect on performance across all evaluated datasets. However, it leads to an increase in inference latency, which is attributable to the additional computational overhead required by the SAM model to process a larger number of points. In contrast, decreasing the sampling density provides only a marginal reduction in latency while inducing a slight performance degradation. Therefore, we adopt the settings $m_p=10, g_p=5$ as they provide an optimal balance between model performance and computational efficiency.

Table~\ref{tab:lambda_mse} illustrates the effect of the hyperparameter $\lambda_{mse}$, which governs the trade-off between $\mathcal{L}_{mse}$ and $\mathcal{L}_{ce}^{mask}$. We report the mIoU of the COCO-Stuff validation set and the final $\mathcal{L}_{mse}$ loss. The findings suggest that $\lambda_{mse}=0.5$ is an appropriate weighting for $\mathcal{L}_{mse}$. Moreover, a higher $\lambda_{mse}$ instead leads to an increase in $\mathcal{L}_{mse}$, as a higher weight on $\mathcal{L}_{mse}$ can detrimentally impact segmentation performance, which is necessary for probability learning.

\begin{table}[htbp]
\centering
\vspace{-7pt}
\caption{The impact of the hyperparameters on TSPP sampling density. $m_p$ and $g_p$ are hyperparameters that control the density of points sampled by TSPP.}
\label{tab:tspp_hyper}
\begin{tabular}{c|c|ccccc}
\toprule
Hyperparameters & Latency$\downarrow$ & A-847 & PC-459 & A-150 & PC-59 & PAS-20 \\ \midrule
$m_p$=12, $g_p$=5 & 1082    & 12.5  & 19.8   & 32.9  & 58.5  & 95.2   \\
$m_p$=10, $g_p$=4 & 994     & 12.5  & 19.8   & 32.9  & 58.4  & 95.2   \\ \midrule
$m_p$=\underline{10}, $g_p$=\underline{5} & 950     & 12.5  & 19.8   & 32.9  & 58.4  & 95.2   \\ \midrule
$m_p$=10, $g_p$=6 & 945     & 12.3  & 19.7   & 32.8  & 58.4  & 95.1   \\
$m_p$=8, $g_p$=5  & 937     & 12.4  & 19.6   & 32.8  & 58.3  & 95.1   \\ \bottomrule
\end{tabular}
\vspace{-7pt}
\end{table}

\begin{table}[htbp]
\centering
\caption{Impact of hyperparameters on training TSPP. $\lambda_{mse}$ is the hyperparameter to balance the multiple losses in TSPP.}
\begin{tabular}{l|cc}
\toprule
$\lambda_{mse}$ & mIoU $\uparrow$      & $\mathcal{L}_{mse} (10^{-3})$ $\downarrow$\\ \midrule
0.1             & 30.7              & 8.0                           \\
\underline{0.5}             & \textbf{31.1}              & \textbf{7.8}                           \\
1.0               & 30.5              & 15.1                          \\ \bottomrule
\end{tabular}
\label{tab:lambda_mse}
\end{table}

\noindent \textbf{Impact of different SAM variants.}
Table~\ref{tab:sam} presents the impact of employing various SAM variants. These models are grouped into two categories. \textbf{(1) Variants designed to enhance segmentation capability, including HQ-SAM~\cite{ke2024segment} and SAM2~\cite{ravi2024sam}.} Similar to the original SAM-H, we utilize the largest available pre-trained models, HQ-SAM-H and SAM2-L. Our findings indicate that the performance of HQ-SAM is nearly identical to that of the original SAM. This similarity can be attributed to the current grid-based point prompts used to extract all objects. Each point generates a single mask, and this non-interactive prompting strategy does not fully leverage the advantages of HQ-SAM, which requires specific interactive cues (such as clicking on challenging regions) to refine complex boundaries and intricate structures. Moreover, using SAM2-L results in a slight performance degradation. We find that this is because SAM2's primary enhancements are for video processing, which may have come at the cost of image segmentation quality. We observed that in its segment everything mode, SAM2 fails to segment many objects because the corresponding masks are of low quality (e.g., containing significant edge noise) and are therefore discarded. \textbf{(2) Variants optimized for inference speed, including FastSAM~\cite{zhao2023fast} and MobileSAM~\cite{zhang2023faster}.} These models achieve a significant latency reduction of approximately 300 ms and a significant decrease in parameter count. However, this gain in efficiency is achieved at the expense of a loss in segmentation accuracy, with a consistent performance drop observed across all datasets.

\begin{table}[]
\centering
\caption{Impact of Different SAM Variants. Comparison of original SAM with improved variants (HQ-SAM, SAM2) and fast SAM (FastSAM, MobileSAM).}
\label{tab:sam}
\resizebox{\textwidth}{!}{
\begin{tabular}{l|cc|ccccc}
\toprule
SAM Type     & Params (M)  & Latency$\downarrow$  & A-847 & PC-459 & A-150 & PC-59 & PAS-20 \\ \midrule
Original SAM-H~\cite{kirillov2023segment} & 641    & 950     & 12.5  & 19.8   & 32.9  & 58.4  & 95.2   \\ \midrule
HQ-SAM-H~\cite{ke2024segment}       & 612    & 1099    & 12.5  & 19.9   & 32.8  & 58.5  & 95.1   \\ 
SAM2-L~\cite{ravi2024sam}         & 214    & 760     & 12.3  & 19.5   & 32.6  & 58.2  & 95.1   \\ \midrule
FastSAM~\cite{zhao2023fast}      & 70     & 675     & 12.4  & 19.6   & 32.7  & 58.3  & 95.0   \\
MobileSAM~\cite{zhang2023faster}    & 10     & 661     & 12.3  & 19.5   & 32.5  & 58.1  & 94.9   \\ \bottomrule
\end{tabular}
}
\end{table}

\noindent \textbf{Model efficiency analysis.}
We analyze the model's efficiency from both training and inference perspectives. Table~\ref{tab:efficiency} demonstrates the training efficiency of SAM-MI. Despite incorporating SAM's substantial number of parameters (641M frozen parameters), our integration strategy maintains practical applicability by limiting the increase in trainable parameters (+3M) and the increase in training latency (+275ms) compared to baseline models. xFurthermore, we evaluate the inference performance against Grounded-SAM, as detailed in Table~\ref {tab:inference}. While the total number of parameter are comparable, SAM-MI is more streamlined with 35M fewer parameters. More significantly, our method achieves a substantial 1.6$\times$ speedup in inference latency. This considerable acceleration is primarily attributable to our proposed TSPP sparse sampling strategy, highlighting its effectiveness in optimizing inference speed.

\begin{table}[h!]
    \centering
    \caption{Model training efficiency analysis.}
    \label{tab:efficiency}
    \begin{tabular}{lccc}
    \toprule
    \multirow{2}{*}{SAM-MI} & \multirow{2}{*}{Train Latency (ms/iter)} & \multicolumn{2}{c}{Params (M)}         \\
    \cmidrule(lr){3-4}
                            &                                       & Trainable & \multicolumn{1}{l}{Total} \\ \midrule
    $w/o$ SAM                 & 471                                   & 25  & 154  \\
    $w/$ SAM                  & 746                                    & 28    &798  \\ \bottomrule
    \end{tabular}
\end{table}

\begin{table}[h!]
\centering
\caption{Model inference efficiency analysis.}
\label{tab:inference}
\begin{tabular}{l|ccc}
\toprule
Method       & Params (M) & Inference Latency (ms) & MESS Avg. mIoU      \\ \midrule
Grounded-SAM & 833        & 1537                        & 28.52     \\
SAM-MI       & 798        & 950                         & 33.28     \\ \bottomrule
\end{tabular}
\end{table}

\begin{figure}[t]
    \centering
    \includegraphics[width=\linewidth]{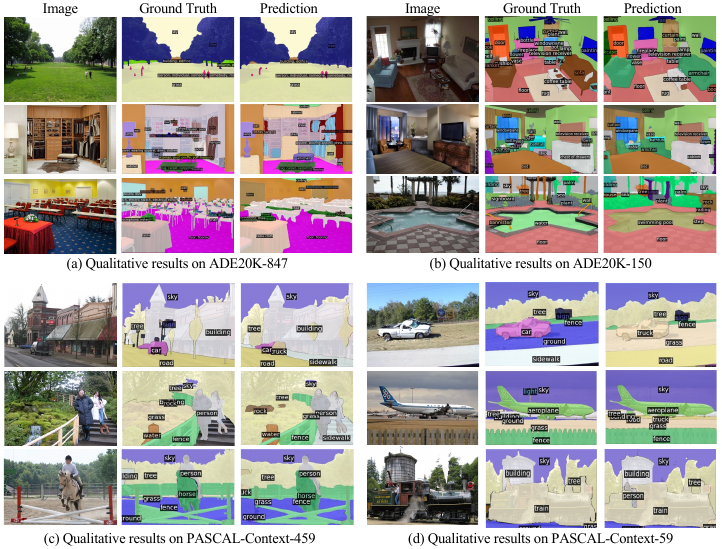}
    \caption{Visualization of segmentation results on ADE20K-847, ADE20K-150, PASCAL-Context-459 and PASCAL-Context-59.}
    \label{fig:more-quality}
\end{figure}

\noindent \textbf{Failure case analysis.}
Fig.~\ref{fig:more-quality} presents more qualitative results of our SAM-MI. We identify two primary failure modes. \textbf{(1) Extremely small and narrow objects.} While our SAM-MI demonstrates the ability to segment and recognize small objects, as evidenced by its capability to identify tiny people in Fig.~\ref{fig:more-quality} (a), it still encounters significant challenges when dealing with extremely small or narrow objects, such as chair legs in Fig.~\ref{fig:more-quality} (a), fan in Fig.~\ref{fig:more-quality} (b) and trees in Fig.~\ref{fig:more-quality} (c). There are two reasons for this: (i) Sparse sampling points may not adequately cover small objects. This problem can be alleviated by increasing the number of sampling points, but at the expense of decreasing inference speed. (ii) Even if these small objects are segmented, CLIP exhibits poor recognition ability for small objects due to the lack of pixel-level representation capabilities. This deficiency necessitates further investigation. In addition, the difficulty of small object segmentation is also prominent for other existing methods. \textbf{(2) Extremely complex backgrounds.} Our method demonstrates a commendable ability to segment objects within moderately complex backgrounds, as seen in its capacity to handle the indoor scene with many classes in Fig.~\ref{fig:more-quality} (b) and the intricate outdoor environment with overlapping trees and buildings in Fig.~\ref{fig:more-quality} (c). However, the model still struggles with scenarios characterized by an extremely complex background, such as the model failing to recognize the shoes and clothes from the furniture. It is important to note that these challenges are not exclusive to our model. The accurate segmentation of small objects and effective disentanglement of objects in complex background scenes are ongoing and unresolved issues in the broader field of semantic segmentation.

Nevertheless, we notice that these datasets for the current OVSS evaluation exhibit certain shortcomings: (1) numerous objects lack proper annotations (e.g., the unannotated person in Fig.~\ref{fig:more-quality} (d)) or have incorrect labels (e.g., the mislabeled swimming pool in Fig.~\ref{fig:more-quality} (b)), and (2) there are non-independent category definitions. For instance, A-847 defines animals like dogs, cats, and birds while also including the animal class. As a result, there is a pressing need to establish a reliable benchmark for evaluating OVSS models.

\section{Conclusion}
\label{sec:conclusion}
In this paper, we present SAM-MI, a novel mask-injected framework for equipping open-vocabulary semantic segmentation (OVSS) models with the universal segmentation capability from SAM. SAM-MI introduces Shallow Mask Aggregation (SMAgg), Decoupled Mask Injection (DMI), and Text-guided Sparse Point Prompter (TSPP) in a unified framework. Considering the tending of SAM to segment numerous out-of-interest masks with colour clues, SMAgg models the relationship between masks to reassemble partial over-segmented masks. In contrast to previous methodologies that associate fixed SAM-generated masks with labels, DMI only employs them as guidance to enhance the contextual information, making it robust to inaccurate masks. Additionally, by probabilistic sampling, TSPP avoids dense grid-like point prompts, significantly reducing the computational overhead. Our proposed SAM-MI outperforms existing methods across diverse datasets, demonstrating that it is a practical methodology for incorporating OVSS with SAM.

\noindent \textbf{Limitations and future work.}
As indicated by our failure case analysis, a significant limitation lies in the current benchmarks for OVSS evaluation: poorly modeled class hierarchies and prevalence with incomplete or incorrect annotations. Establishing a more reliable and rigorous benchmark is thus a critical need for the field. Furthermore, while the study primarily focuses on semantic segmentation, extending our proposed mask-injected framework to other dense prediction tasks, such as instance or panoptic segmentation, presents a promising avenue for future work.

\section{Acknowledgements}
This research was supported by the National Natural Science Foundation of China under Grant 62306310.

\section{Declarations of conflict of interest}
The authors declare that they have no conflicts of interest to this work.


\setcounter{section}{0}
\renewcommand{\thesection}{\Alph{section}}


\begin{thebibliography}{10}
\providecommand{\url}[1]{#1}
\csname url@samestyle\endcsname
\providecommand{\newblock}{\relax}
\providecommand{\bibinfo}[2]{#2}
\providecommand{\BIBentrySTDinterwordspacing}{\spaceskip=0pt\relax}
\providecommand{\BIBentryALTinterwordstretchfactor}{4}
\providecommand{\BIBentryALTinterwordspacing}{\spaceskip=\fontdimen2\font plus
\BIBentryALTinterwordstretchfactor\fontdimen3\font minus \fontdimen4\font\relax}
\providecommand{\BIBforeignlanguage}[2]{{%
\expandafter\ifx\csname l@#1\endcsname\relax
\typeout{** WARNING: IEEEtran.bst: No hyphenation pattern has been}%
\typeout{** loaded for the language `#1'. Using the pattern for}%
\typeout{** the default language instead.}%
\else
\language=\csname l@#1\endcsname
\fi
#2}}
\providecommand{\BIBdecl}{\relax}
\BIBdecl

\bibitem{xu2022simple}
M.~Xu, Z.~Zhang, F.~Wei, Y.~Lin, Y.~Cao, H.~Hu, and X.~Bai, ``A simple baseline for open-vocabulary semantic segmentation with pre-trained vision-language model,'' in \emph{Proceedings of the European Conference on Computer Vision}, 2022, pp. 736--753.

\bibitem{ding2022decoupling}
J.~Ding, N.~Xue, G.-S. Xia, and D.~Dai, ``Decoupling zero-shot semantic segmentation,'' in \emph{Proceedings of the IEEE/CVF Conference on Computer Vision and Pattern Recognition}, 2022, pp. 11\,583--11\,592.

\bibitem{liang2023open}
F.~Liang, B.~Wu, X.~Dai, K.~Li, Y.~Zhao, H.~Zhang, P.~Zhang, P.~Vajda, and D.~Marculescu, ``Open-vocabulary semantic segmentation with mask-adapted clip,'' in \emph{Proceedings of the IEEE/CVF Conference on Computer Vision and Pattern Recognition}, 2023, pp. 7061--7070.

\bibitem{dong2023maskclip}
X.~Dong, J.~Bao, Y.~Zheng, T.~Zhang, D.~Chen, H.~Yang, M.~Zeng, W.~Zhang, L.~Yuan, D.~Chen \emph{et~al.}, ``Maskclip: Masked self-distillation advances contrastive language-image pretraining,'' in \emph{Proceedings of the IEEE/CVF Conference on Computer Vision and Pattern Recognition}, 2023, pp. 10\,995--11\,005.

\bibitem{xu2023open}
J.~Xu, S.~Liu, A.~Vahdat, W.~Byeon, X.~Wang, and S.~De~Mello, ``Open-vocabulary panoptic segmentation with text-to-image diffusion models,'' in \emph{Proceedings of the IEEE/CVF Conference on Computer Vision and Pattern Recognition}, 2023, pp. 2955--2966.

\bibitem{xu2023side}
M.~Xu, Z.~Zhang, F.~Wei, H.~Hu, and X.~Bai, ``Side adapter network for open-vocabulary semantic segmentation,'' in \emph{Proceedings of the IEEE/CVF Conference on Computer Vision and Pattern Recognition}, 2023, pp. 2945--2954.

\bibitem{yu2024convolutions}
Q.~Yu, J.~He, X.~Deng, X.~Shen, and L.-C. Chen, ``Convolutions die hard: Open-vocabulary segmentation with single frozen convolutional clip,'' \emph{Advances in Neural Information Processing Systems}, vol.~36, 2024.

\bibitem{kirillov2023segment}
A.~Kirillov, E.~Mintun, N.~Ravi, H.~Mao, C.~Rolland, L.~Gustafson, T.~Xiao, S.~Whitehead, A.~C. Berg, W.-Y. Lo \emph{et~al.}, ``Segment anything,'' in \emph{Proceedings of the IEEE/CVF International Conference on Computer Vision}, 2023, pp. 4015--4026.

\bibitem{shan2024open}
X.~Shan, D.~Wu, G.~Zhu, Y.~Shao, N.~Sang, and C.~Gao, ``Open-vocabulary semantic segmentation with image embedding balancing,'' in \emph{Proceedings of the IEEE/CVF Conference on Computer Vision and Pattern Recognition}, 2024, pp. 28\,412--28\,421.

\bibitem{chen2024frozenseg}
X.~Chen, H.~Yang, S.~Jin, X.~Zhu, and H.~Yao, ``Frozenseg: Harmonizing frozen foundation models for open-vocabulary segmentation,'' \emph{arXiv preprint arXiv:2409.03525}, 2024.

\bibitem{chen2023semantic}
J.~Chen, Z.~Yang, and L.~Zhang, ``Semantic segment anything,'' \url{https://github.com/fudan-zvg/Semantic-Segment-Anything}, 2023.

\bibitem{chen2024sam}
P.~Chen, L.~Xie, X.~Huo, X.~Yu, X.~Zhang, Y.~Sun, Z.~Han, and Q.~Tian, ``Sam-cp: Marrying sam with composable prompts for versatile segmentation,'' \emph{arXiv preprint arXiv:2407.16682}, 2024.

\bibitem{wang2024sam}
H.~Wang, P.~K.~A. Vasu, F.~Faghri, R.~Vemulapalli, M.~Farajtabar, S.~Mehta, M.~Rastegari, O.~Tuzel, and H.~Pouransari, ``Sam-clip: Merging vision foundation models towards semantic and spatial understanding,'' in \emph{Proceedings of the IEEE/CVF Conference on Computer Vision and Pattern Recognition}, 2024, pp. 3635--3647.

\bibitem{yuan2024open}
H.~Yuan, X.~Li, C.~Zhou, Y.~Li, K.~Chen, and C.~C. Loy, ``Open-vocabulary sam: Segment and recognize twenty-thousand classes interactively,'' \emph{arXiv preprint arXiv:2401.02955}, 2024.

\bibitem{ren2024grounded}
T.~Ren, S.~Liu, A.~Zeng, J.~Lin, K.~Li, H.~Cao, J.~Chen, X.~Huang, Y.~Chen, F.~Yan \emph{et~al.}, ``Grounded sam: Assembling open-world models for diverse visual tasks,'' \emph{arXiv preprint arXiv:2401.14159}, 2024.

\bibitem{huang2024learning}
J.~Huang, K.~Jiang, J.~Zhang, H.~Qiu, L.~Lu, S.~Lu, and E.~Xing, ``Learning to prompt segment anything models,'' \emph{arXiv preprint arXiv:2401.04651}, 2024.

\bibitem{han2023boosting}
X.~Han, L.~Wei, X.~Yu, Z.~Dou, X.~He, K.~Wang, Z.~Han, and Q.~Tian, ``Boosting segment anything model towards open-vocabulary learning,'' \emph{arXiv preprint arXiv:2312.03628}, 2023.

\bibitem{wang2024open}
Z.~Wang, X.~Xia, Z.~Chen, X.~He, Y.~Guo, M.~Gong, and T.~Liu, ``Open-vocabulary segmentation with unpaired mask-text supervision,'' \emph{arXiv preprint arXiv:2402.08960}, 2024.

\bibitem{zhou2019semantic}
B.~Zhou, H.~Zhao, X.~Puig, T.~Xiao, S.~Fidler, A.~Barriuso, and A.~Torralba, ``Semantic understanding of scenes through the ade20k dataset,'' \emph{International Journal of Computer Vision}, vol. 127, pp. 302--321, 2019.

\bibitem{mottaghi2014role}
R.~Mottaghi, X.~Chen, X.~Liu, N.-G. Cho, S.-W. Lee, S.~Fidler, R.~Urtasun, and A.~Yuille, ``The role of context for object detection and semantic segmentation in the wild,'' in \emph{Proceedings of the IEEE/CVF Conference on Computer Vision and Pattern Recognition}, 2014, pp. 891--898.

\bibitem{MESSBenchmark2023}
B.~Blumenstiel, J.~Jakubik, H.~Kühne, and M.~Vössing, ``{What a MESS: Multi-Domain Evaluation of Zero-shot Semantic Segmentation},'' \emph{Advances in Neural Information Processing Systems}, 2023.

\bibitem{li2023semantic}
F.~Li, H.~Zhang, P.~Sun, X.~Zou, S.~Liu, J.~Yang, C.~Li, L.~Zhang, and J.~Gao, ``Semantic-sam: Segment and recognize anything at any granularity,'' \emph{arXiv preprint arXiv:2307.04767}, 2023.

\bibitem{zou2024segment}
X.~Zou, J.~Yang, H.~Zhang, F.~Li, L.~Li, J.~Wang, L.~Wang, J.~Gao, and Y.~J. Lee, ``Segment everything everywhere all at once,'' \emph{Advances in Neural Information Processing Systems}, vol.~36, 2024.

\bibitem{ke2024segment}
L.~Ke, M.~Ye, M.~Danelljan, Y.-W. Tai, C.-K. Tang, F.~Yu \emph{et~al.}, ``Segment anything in high quality,'' \emph{Advances in Neural Information Processing Systems}, vol.~36, 2024.

\bibitem{zhao2023fast}
X.~Zhao, W.~Ding, Y.~An, Y.~Du, T.~Yu, M.~Li, M.~Tang, and J.~Wang, ``Fast segment anything,'' \emph{arXiv preprint arXiv:2306.12156}, 2023.

\bibitem{zhang2023faster}
C.~Zhang, D.~Han, Y.~Qiao, J.~U. Kim, S.-H. Bae, S.~Lee, and C.~S. Hong, ``Faster segment anything: Towards lightweight sam for mobile applications,'' \emph{arXiv preprint arXiv:2306.14289}, 2023.

\bibitem{xiong2024efficientsam}
Y.~Xiong, B.~Varadarajan, L.~Wu, X.~Xiang, F.~Xiao, C.~Zhu, X.~Dai, D.~Wang, F.~Sun, F.~Iandola \emph{et~al.}, ``Efficientsam: Leveraged masked image pretraining for efficient segment anything,'' in \emph{Proceedings of the IEEE/CVF Conference on Computer Vision and Pattern Recognition}, 2024, pp. 16\,111--16\,121.

\bibitem{ji2024segment}
W.~Ji, J.~Li, Q.~Bi, T.~Liu, W.~Li, and L.~Cheng, ``Segment anything is not always perfect: An investigation of sam on different real-world applications,'' 2024.

\bibitem{liu2025balanced}
H.~Liu, Y.~Wang, M.~Ren, J.~Hu, Z.~Luo, G.~Hou, and Z.~Sun, ``Balanced representation learning for long-tailed skeleton-based action recognition,'' \emph{Machine Intelligence Research}, pp. 1--18, 2025.

\bibitem{wang2023caption}
T.~Wang, J.~Zhang, J.~Fei, H.~Zheng, Y.~Tang, Z.~Li, M.~Gao, and S.~Zhao, ``Caption anything: Interactive image description with diverse multimodal controls,'' \emph{arXiv preprint arXiv:2305.02677}, 2023.

\bibitem{gao2023editanything}
S.~Gao, Z.~Lin, X.~Xie, P.~Zhou, M.-M. Cheng, and S.~Yan, ``Editanything: Empowering unparalleled flexibility in image editing and generation,'' in \emph{Proceedings of the 31st ACM International Conference on Multimedia, Demo track}, 2023.

\bibitem{li2024clipsam}
S.~Li, J.~Cao, P.~Ye, Y.~Ding, C.~Tu, and T.~Chen, ``Clipsam: Clip and sam collaboration for zero-shot anomaly segmentation,'' \emph{arXiv preprint arXiv:2401.12665}, 2024.

\bibitem{cao2023segment}
Y.~Cao, X.~Xu, C.~Sun, Y.~Cheng, Z.~Du, L.~Gao, and W.~Shen, ``Segment any anomaly without training via hybrid prompt regularization,'' \emph{arXiv preprint arXiv:2305.10724}, 2023.

\bibitem{liu2024deep}
J.~Liu, G.~Xie, J.~Wang, S.~Li, C.~Wang, F.~Zheng, and Y.~Jin, ``Deep industrial image anomaly detection: A survey,'' \emph{Machine Intelligence Research}, vol.~21, no.~1, pp. 104--135, 2024.

\bibitem{zhu2024medical}
J.~Zhu, Y.~Qi, and J.~Wu, ``Medical sam 2: Segment medical images as video via segment anything model 2,'' \emph{arXiv preprint arXiv:2408.00874}, 2024.

\bibitem{ma2024segment}
J.~Ma, Y.~He, F.~Li, L.~Han, C.~You, and B.~Wang, ``Segment anything in medical images,'' \emph{Nature Communications}, vol.~15, no.~1, p. 654, 2024.

\bibitem{radford2021learning}
A.~Radford, J.~W. Kim, C.~Hallacy, A.~Ramesh, G.~Goh, S.~Agarwal, G.~Sastry, A.~Askell, P.~Mishkin, J.~Clark \emph{et~al.}, ``Learning transferable visual models from natural language supervision,'' in \emph{International Conference on Machine Learning}, 2021, pp. 8748--8763.

\bibitem{jia2021scaling}
C.~Jia, Y.~Yang, Y.~Xia, Y.-T. Chen, Z.~Parekh, H.~Pham, Q.~Le, Y.-H. Sung, Z.~Li, and T.~Duerig, ``Scaling up visual and vision-language representation learning with noisy text supervision,'' in \emph{International Conference on Machine Learning}, 2021, pp. 4904--4916.

\bibitem{li2022blip}
J.~Li, D.~Li, C.~Xiong, and S.~Hoi, ``Blip: Bootstrapping language-image pre-training for unified vision-language understanding and generation,'' in \emph{International Conference on Machine Learning}, 2022, pp. 12\,888--12\,900.

\bibitem{zhong2022regionclip}
Y.~Zhong, J.~Yang, P.~Zhang, C.~Li, N.~Codella, L.~H. Li, L.~Zhou, X.~Dai, L.~Yuan, Y.~Li \emph{et~al.}, ``Regionclip: Region-based language-image pretraining,'' in \emph{Proceedings of the IEEE/CVF Conference on Computer Vision and Pattern Recognition}, 2022, pp. 16\,793--16\,803.

\bibitem{zhu2024mrovseg}
Y.~Zhu, B.~Zhu, Z.~Chen, H.~Xu, M.~Tang, and J.~Wang, ``Mrovseg: Breaking the resolution curse of vision-language models in open-vocabulary semantic segmentation,'' \emph{arXiv preprint arXiv:2408.14776}, 2024.

\bibitem{cho2024cat}
S.~Cho, H.~Shin, S.~Hong, A.~Arnab, P.~H. Seo, and S.~Kim, ``Cat-seg: Cost aggregation for open-vocabulary semantic segmentation,'' in \emph{Proceedings of the IEEE/CVF Conference on Computer Vision and Pattern Recognition}, 2024, pp. 4113--4123.

\bibitem{xie2024sed}
B.~Xie, J.~Cao, J.~Xie, F.~S. Khan, and Y.~Pang, ``Sed: A simple encoder-decoder for open-vocabulary semantic segmentation,'' in \emph{Proceedings of the IEEE/CVF Conference on Computer Vision and Pattern Recognition}, 2024, pp. 3426--3436.

\bibitem{chen2025errseg}
Lin Chen, Qi~Yang, Kun Ding, Zhihao Li, Gang Shen, Fei Li, Qiyuan Cao, and Shiming Xiang.
\newblock Efficient redundancy reduction for open-vocabulary semantic segmentation.
\newblock {\em arXiv preprint arXiv:2501.17642}, 2025.

\bibitem{long2015fully}
J.~Long, E.~Shelhamer, and T.~Darrell, ``Fully convolutional networks for semantic segmentation,'' in \emph{Proceedings of the IEEE/CVF Conference on Computer Vision and Pattern Recognition}, 2015, pp. 3431--3440.

\bibitem{liu2023grounding}
S.~Liu, Z.~Zeng, T.~Ren, F.~Li, H.~Zhang, J.~Yang, C.~Li, J.~Yang, H.~Su, J.~Zhu \emph{et~al.}, ``Grounding dino: Marrying dino with grounded pre-training for open-set object detection,'' \emph{arXiv preprint arXiv:2303.05499}, 2023.

\bibitem{zhang2023personalize}
R.~Zhang, Z.~Jiang, Z.~Guo, S.~Yan, J.~Pan, X.~Ma, H.~Dong, P.~Gao, and H.~Li, ``Personalize segment anything model with one shot,'' \emph{arXiv preprint arXiv:2305.03048}, 2023.

\bibitem{gaillochet2024automating}
M.~Gaillochet, C.~Desrosiers, and H.~Lombaert, ``Automating medsam by learning prompts with weak few-shot supervision,'' in \emph{International Workshop on Foundation Models for General Medical AI}.\hskip 1em plus 0.5em minus 0.4em\relax Springer, 2024, pp. 61--70.

\bibitem{rao2022denseclip}
Y.~Rao, W.~Zhao, G.~Chen, Y.~Tang, Z.~Zhu, G.~Huang, J.~Zhou, and J.~Lu, ``Denseclip: Language-guided dense prediction with context-aware prompting,'' in \emph{Proceedings of the IEEE/CVF Conference on Computer Vision and Pattern Recognition}, 2022, pp. 18\,082--18\,091.

\bibitem{caesar2018coco}
H.~Caesar, J.~Uijlings, and V.~Ferrari, ``Coco-stuff: Thing and stuff classes in context,'' in \emph{Proceedings of the IEEE/CVF Conference on Computer Vision and Pattern Recognition}, 2018, pp. 1209--1218.

\bibitem{everingham2010pascal}
M.~Everingham, L.~Van~Gool, C.~K. Williams, J.~Winn, and A.~Zisserman, ``The pascal visual object classes (voc) challenge,'' \emph{International Journal of Computer Vision}, vol.~88, pp. 303--338, 2010.

\bibitem{yu2020bdd100kdiversedrivingdataset}
\BIBentryALTinterwordspacing
F.~Yu, H.~Chen, X.~Wang, W.~Xian, Y.~Chen, F.~Liu, V.~Madhavan, and T.~Darrell, ``Bdd100k: A diverse driving dataset for heterogeneous multitask learning,'' \emph{arXiv preprint arXiv:1805.04687}, 2020.
\BIBentrySTDinterwordspacing

\bibitem{waqas2019isaid}
S.~Waqas~Zamir, A.~Arora, A.~Gupta, S.~Khan, G.~Sun, F.~Shahbaz~Khan, F.~Zhu, L.~Shao, G.-S. Xia, and X.~Bai, ``isaid: A large-scale dataset for instance segmentation in aerial images,'' in \emph{Proceedings of the IEEE/CVF Conference on Computer Vision and Pattern Recognition Workshops}, 2019, pp. 28--37.

\bibitem{mahbod2021cryonuseg}
A.~Mahbod, G.~Schaefer, B.~Bancher, C.~L{\"o}w, G.~Dorffner, R.~Ecker, and I.~Ellinger, ``Cryonuseg: A dataset for nuclei instance segmentation of cryosectioned h\&e-stained histological images,'' \emph{Computers in Biology and Medicine}, vol. 132, p. 104349, 2021.

\bibitem{zou2018deepcrack}
Q.~Zou, Z.~Zhang, Q.~Li, X.~Qi, Q.~Wang, and S.~Wang, ``Deepcrack: Learning hierarchical convolutional features for crack detection,'' \emph{IEEE Transactions on Image Processing}, vol.~28, no.~3, pp. 1498--1512, 2018.

\bibitem{haug2015crop}
S.~Haug and J.~Ostermann, ``A crop/weed field image dataset for the evaluation of computer vision based precision agriculture tasks,'' in \emph{Computer Vision-ECCV 2014 Workshops: Zurich, Switzerland, September 6-7 and 12, 2014, Proceedings, Part IV 13}.\hskip 1em plus 0.5em minus 0.4em\relax Springer, 2015, pp. 105--116.

\bibitem{cherti2023reproducible}
M.~Cherti, R.~Beaumont, R.~Wightman, M.~Wortsman, G.~Ilharco, C.~Gordon, C.~Schuhmann, L.~Schmidt, and J.~Jitsev, ``Reproducible scaling laws for contrastive language-image learning,'' in \emph{Proceedings of the IEEE/CVF Conference on Computer Vision and Pattern Recognition}, 2023, pp. 2818--2829.

\bibitem{dosovitskiy2020vit}
A.~Dosovitskiy, L.~Beyer, A.~Kolesnikov, D.~Weissenborn, X.~Zhai, T.~Unterthiner, M.~Dehghani, M.~Minderer, G.~Heigold, S.~Gelly, J.~Uszkoreit, and N.~Houlsby, ``An image is worth 16x16 words: Transformers for image recognition at scale,'' \emph{ICLR}, 2021.

\bibitem{loshchilov2017decoupled}
I.~Loshchilov and F.~Hutter, ``Decoupled weight decay regularization,'' \emph{arXiv preprint arXiv:1711.05101}, 2017.

\bibitem{zou2023generalized}
X.~Zou, Z.-Y. Dou, J.~Yang, Z.~Gan, L.~Li, C.~Li, X.~Dai, H.~Behl, J.~Wang, L.~Yuan \emph{et~al.}, ``Generalized decoding for pixel, image, and language,'' in \emph{Proceedings of the IEEE/CVF Conference on Computer Vision and Pattern Recognition}, 2023, pp. 15\,116--15\,127.

\bibitem{zhang2023simple}
H.~Zhang, F.~Li, X.~Zou, S.~Liu, C.~Li, J.~Yang, and L.~Zhang, ``A simple framework for open-vocabulary segmentation and detection,'' in \emph{Proceedings of the IEEE/CVF International Conference on Computer Vision}, 2023, pp. 1020--1031.

\bibitem{han2023open}
C.~Han, Y.~Zhong, D.~Li, K.~Han, and L.~Ma, ``Open-vocabulary semantic segmentation with decoupled one-pass network,'' in \emph{Proceedings of the IEEE/CVF International Conference on Computer Vision}, 2023, pp. 1086--1096.

\bibitem{liu2024open}
Y.~Liu, S.~Bai, G.~Li, Y.~Wang, and Y.~Tang, ``Open-vocabulary segmentation with semantic-assisted calibration,'' in \emph{Proceedings of the IEEE/CVF Conference on Computer Vision and Pattern Recognition}, 2024, pp. 3491--3500.

\bibitem{jiao2023maft}
S.~Jiao, Y.~Wei, Y.~Wang, Y.~Zhao, and H.~Shi, ``Learning mask-aware clip representations for zero-shot segmentation,'' \emph{Advances in Neural Information Processing Systems}, vol.~36, pp. 35\,631--35\,653, 2023.

\bibitem{jiao2024maftp}
S.~Jiao, H.~Zhu, J.~Huang, Y.~Zhao, Y.~Wei, and H.~Shi, ``Collaborative vision-text representation optimizing for open-vocabulary segmentation,'' in \emph{European Conference on Computer Vision}.\hskip 1em plus 0.5em minus 0.4em\relax Springer, 2024, pp. 399--416.

\bibitem{niu2025eov}
H.~Niu, J.~Hu, J.~Lin, G.~Jiang, and S.~Zhang, ``Eov-seg: Efficient open-vocabulary panoptic segmentation,'' in \emph{Proceedings of the AAAI Conference on Artificial Intelligence}, vol.~39, no.~6, 2025, pp. 6254--6262.

\bibitem{zeng2024maskclippp}
Q.-S. Zeng, Y.~Li, D.~Zhou, G.~Li, Q.~Hou, and M.-M. Cheng, ``High-quality mask tuning matters for open-vocabulary segmentation,'' \emph{arXiv preprint arXiv:2412.11464}, 2024.

\bibitem{ravi2024sam}
N.~Ravi, V.~Gabeur, Y.-T. Hu, R.~Hu, C.~Ryali, T.~Ma, H.~Khedr, R.~R{\"a}dle, C.~Rolland, L.~Gustafson \emph{et~al.}, ``Sam 2: Segment anything in images and videos,'' \emph{arXiv preprint arXiv:2408.00714}, 2024.

\end{thebibliography}
\end{document}